# UAVs and Birds: Enhancing Short-Range Navigation through Budgerigar Flight Studies

**Md. Mahmudur Rahman, Sajid Islam, Showren Chowdhury, Sadia Jahan Zeba and Debajyoti Karmaker**
Department of Computer Science
American International University-Bangladesh

**Abstract:**

This study delves into the flight behaviors of Budgerigars (Melopsittacus undulatus) to gain insights into their flight trajectories and movements. Using 3D reconstruction from stereo video camera recordings, we closely examine the velocity and acceleration patterns during three flight motion takeoff, flying and landing. The findings not only contribute to our understanding of bird behaviors but also hold significant implications for the advancement of algorithms in Unmanned Aerial Vehicles (UAVs). The research aims to bridge the gap between biological principles observed in birds and the application of these insights in developing more efficient and autonomous UAVs. In the context of the increasing use of drones, this study focuses on the biologically inspired principles drawn from bird behaviors, particularly during takeoff, flying and landing flight, to enhance UAV capabilities. The dataset created for this research sheds light on Budgerigars' takeoff, flying, and landing techniques, emphasizing their ability to control speed across different situations and surfaces. The study underscores the potential of incorporating these principles into UAV algorithms, addressing challenges related to short-range navigation, takeoff, flying, and landing.

***Keywords:*** *Bird Behaviors, Bio-Inspired Flight, 3D Trajectory Analysis, Takeoff-Flying-Landing Motion analysis*

## 1. Introduction

In recent years there has been a rise in the research area to understand how birds use different techniques while flying to avoid and maneuver through various obstacles and conditions. The idea is to understand their biological features and figure out ways to implement them in developing new and improved algorithms to assist in various applications such as unmanned aerial vehicles (UAV), robotics, autonomous vehicles, and many more. Flying animals are not only incredible navigators but also display impressive dexterity in avoiding collisions and energy preservation techniques. In recent years computer vision techniques have been extensively used in researching bird's and insect's flight motions to extract data and develop complex algorithms to use in UAVs to improve their flights.

Increase in usage of drones has led to vast research being conducted on making these machines fly more smoothly and efficiently. One main sector for such research is biologically inspired principles taken from birds and insects to be implemented on these drones for them to be fully autonomous or worthy of being called unmanned aerial vehicles or UAV.

Research in this area includes tracking bird's motions in controlled environments, finding their landing and takeoff properties, their maneuvering properties in different terrains, collision avoidance and visual guidance. Great interest has been raised in how birds navigate and flies through obstacles. Also, how they can fly robustly in different wind conditions. Another important issue is to concern how they face the challenges to control the speed while taking off and landing in different situations and surface.

In this study we investigate how the birds (Budgerigars) perform takeoff, flying and landing by controlling their speed through one parch to another and mitigate short range navigation. The goal of these investigations is to learn more about the principles of visually directed flight in flying species and these concepts can be helpful for manipulating algorithms for UAV's short range navigation, takeoff, flying and landing.

## 2. Literature Review

Over the past few years, extensive research has been conducted into finding the intricate ways in which birds, despite their small brains, navigate, control their speed, and execute flight strategies. Some notable research on these topics includes how birds employ vision to regulate flight speed within confined spaces, such as narrow tunnels equipped with black and white stripes for precise measurements [1] [2], the narrow tunnel was 25m in length and equipped with black and white stripes to aid with measurements. Utilizing 3D cameras inside these tunnels, researchers have substantiated the hypothesis that birds rely on optical flow-based navigation. Understanding their methodologies gives more knowledge to work on bird flight trajectories.

Schiffner and this team focused on determining a bird's velocity within tunnels of varying widths [3]. Observations revealed that birds switch between high speeds in wider sections of a tunnel and low speeds in the narrower sections. On their paper they measured the birds speed of 9.44m/s in wider sections while on narrower sections the speed dropped to 5.44m/s. On a related paper, birds' ability was explored to avoid obstacle avoiding capabilities and how they strategically bias their direction towards larger visual gaps while making rapid steering decisions [4]. Again, such research about bird trajectories will give us more light onto how to proceed with creating a dataset consisting of bird's flight trajectories.

Bird's avoiding collision is a fascinating phenomenon that researchers have been observing for various reasons. One such reason has captivated researchers, driven in part by the potential application of such insights in Unmanned Aerial Vehicles (UAVs). A particular study shows how birds avoid mid-air collisions during head-on encounters, outlining criteria such as proximity before evasion, directional tendencies, and altitude preferences [5]. Also they describe how they observed two birds flying towards each other in a tunnel to avoid the collision. Their three main criteria for the observations were; 1) how close the encounter is before they evade it; 2) each bird's tendency to go left or right and its consistency; 3) each birds tendency to go above or below the other bird. Their analysis reveals that birds exhibit a preference to veer to the right and maintain a preferred altitude to avoid collisions in head-on encounters, also the collision avoidance behavior does not depend upon the direction of flight, thus ruling out the possible involvement of an internal compass.

One of the study has investigated on bird's landing behavior, which depend on surface properties [6]. Their research shows how birds exhibit stereotypical leg and wing dynamics regardless of perch diameter but how their tow and claw dynamics change depending on the surface properties. Another investigation focuses on how birds use visual edges to target and guide their landing trajectories [7]. They investigated the budgerigar's landings in controlled environments. In their research they used 'concentric circles on a surface with a feeder at the center. The birds would land on the surface with the circle markings in order to gain access to the feeder. They observed that the birds would generally land at the edge of the circles and then stroll to the feeder. This proposes that birds utilize visual differences of the surface they land on to determine where they will land.

Investigation has conduct, the paths taken by budgerigars flying in a tunnel in a controlled environment and then reconstructed their flights in 3D using high speed stereo cameras. Their main goal for the research was to investigate the bird's motion in a controlled environment and to observe their flight patterns in 3D space [8]. This paper has described their approach on how they generated their 3D data to for their computations.

A review paper written by [9] describes a number of the bioinspired research done for UAVs. Their paper focuses on the biologically inspired principles that have been applied on visual guidance, navigation and controlling of UAVs. The paper describes the current limitations on UAVs and how bio-inspired methodologies and principles have and can overcome these limitations. Their main goal is to accumulate all principles for a UAV to become fully autonomous. Birds maneuver through various environments and understanding these principles help in the autonomous movement of UAVs, [10] describes tackling such limitations. Birds offer inspiration on how to fly in these environments and how to exploit complex wind flows while preserving energy. This research highlights how they use GPS backpacks to track birds and their flights in various terrains and extract data to propose a model for optimizing airspeed for wind while maintaining a steady flight trajectory. Similarly, research investigated how birds exploit atmospheric boundary layer in urban environments reducing their energy [11]. This research proposes a method to mimic their movement for a fixed-wing UAV to overcome the energy limitations.

Recognizing the need a dedicated effort created and analyzed a dataset focusing on budgerigars' takeoff, flight, and landing motions. This work also aims to contribute to future computer vision applications in understanding avian behavior.

## 3. Materials and Methods:

### 3.1. Datasets:

For the analysis purpose we have created a dataset. The dataset that consists of the controlled flight of birds with annotated images tracking their motion. 355 clips of birds (Budgerigar) flight motion have been captured in various situations where they show different types of movement and variations speed according to their perspective. The whole process was carried out in an empty enclosed room where we hung two perches where the birds could land and sit. We initially waited for events where the birds would voluntarily fly from one perch to another. We eventually realized that the birds needed to be trained to do this task, so we switched to a more reward system approach to lure the birds with treats to fly from one perch to another.

We used four Budgerigar(budgie), genus Melopsittacus as our main subjects. The four birds were bought from a local breeder and kept indoors in a cage with adequate food and water. From time to time, they were let out inside the enclosed room for recreational purposes. This was necessary so that the birds adapt to flying in the room and get comfortable with the surrounding colors and lights. The room had enough ventilation and an ample amount of airflow for the birds' comfort and the handler. Since the birds are not wild and were bred in captivity, not more than 1 hour was spent each day to use the birds for flight trials. This was done to make sure the birds do not face fatigue.

The recordings were done indoors inside an empty room. The dimensions of the room are 155 inch in length, 120 inch in width, and 118 inch in height. Inside the room, two perches were hung from the ceiling at the height of approximately 84 inches from the floor and 48 to 72-inch distance from each other using GI (Galvanized Iron) wires. The perches were wooden sticks 20-inch-long and around half an inch in diameter. Two cameras were set up at 107 inch in height above the floor on tripods with 70-inch distance between them. The cameras were facing the wall opposite of them and rotated approximately 45 degrees towards each other so that the perches were at the middle of the frame for both cameras. Visualization of the full setup is given in Figure. 1

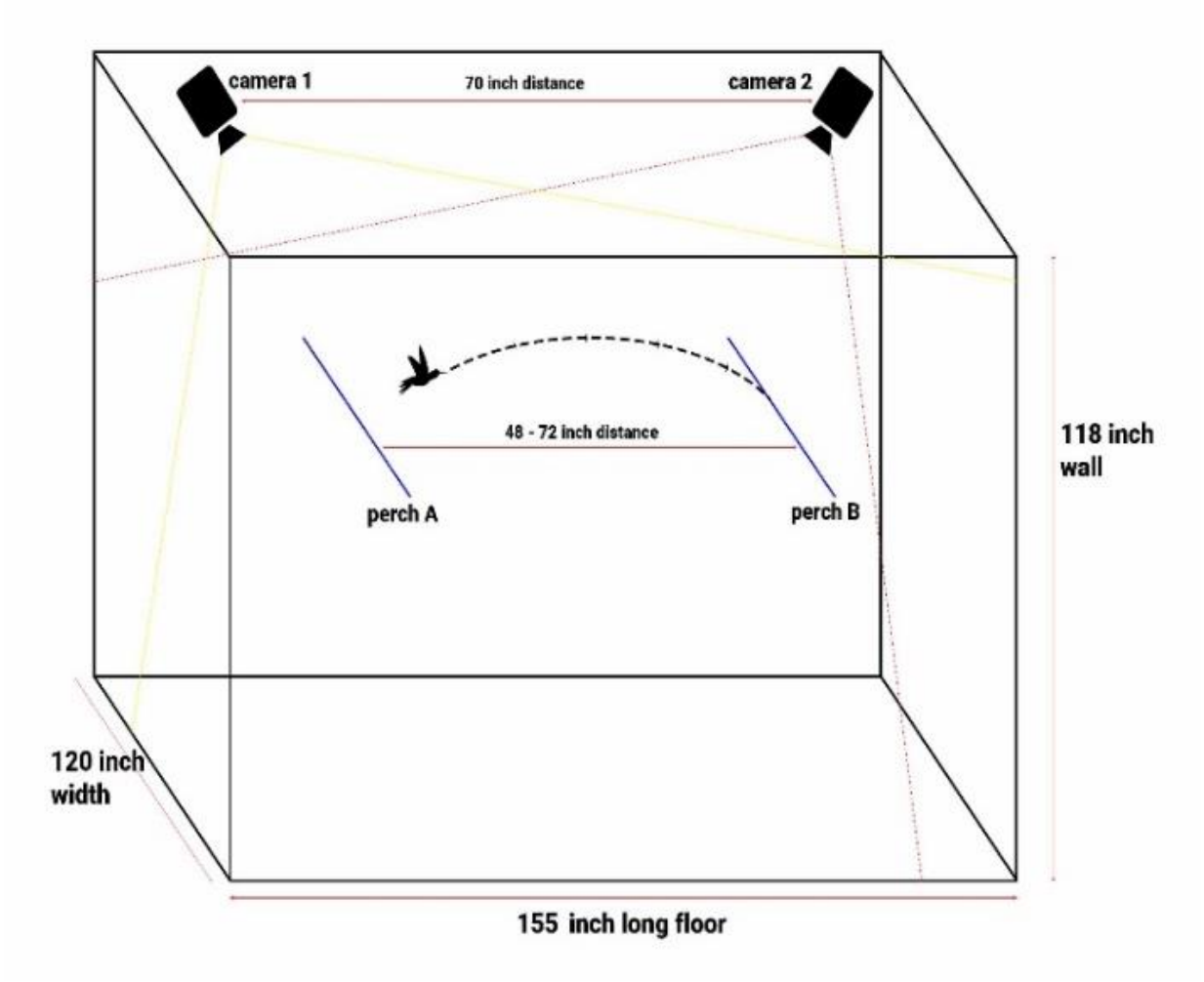

Figure 1. Illustration of the full setup

Two cameras recorded videos at 1920X1080 pixels at 30 frames per second. The cameras were left to record everything from the start of the session till the end. Later the videos were cut into small clips that only captured the desired events. All four birds were left to fly freely in the room during the recording session. They were carefully placed on the perches with a hand-held perch. To trigger an event; a bird flying from one perch to another perch, one of the following methods were followed: 1) Leave the bird to decide for itself and fly to another perch from their sitting perch. This takes time, but it does happen and can be kept in mind during break times and leaving the birds on the perch while the cameras are still recording. 2)Lure the bird with food to the opposite perch. 2)Place the bird on a hand-held perch and twist it or agitate it - this induces the bird to take off and fly to the perch opposite to the handler holding the perch.

The cameras were recording during the entirety of the session to prevent missing any random event. After obtaining the session's video, clips of events from the video were trimmed off and set as singular events for the dataset. Each clip was carefully trimmed from the more extensive video so that each clip has the same set of frames in both the cameras. The clips were then manually annotated using Computer Vision Annotation Tool (CVAT).

For generating the 3D trajectories of the bird's flight motion, mainly Matlab (MathWorks®) was used. Firstly, the cameras were calibrated using the Stereo Camera Calibrator Toolbox for Matlab (MathWorks®). A set of variables (stereo parameter and estimation errors) were obtained to calculate the 3D positions and camera orientation of both camera 1 and 2 relative to each other. We used a reference checkerboard pattern (check size 43.18 mm) for calibration. The mean overall reprojection error in our configuration was estimated to be 0.33 pixels, as reported by the error estimator built into the Stereo Camera Calibrator Toolbox. Upon obtaining the stereo parameters from the calibration, the birds' coordinates were obtained from the annotation JSON file by calculating the midpoints of the bounding boxes. The coordinates of the bird's position from the two cameras were then combined to generate the X, Y and Z coordinates of the bird's position in 3D space. Point clouds were then generated with the 3D coordinates (Figure. 2). Flight paths include a perch to b perch and b perch to a perch.

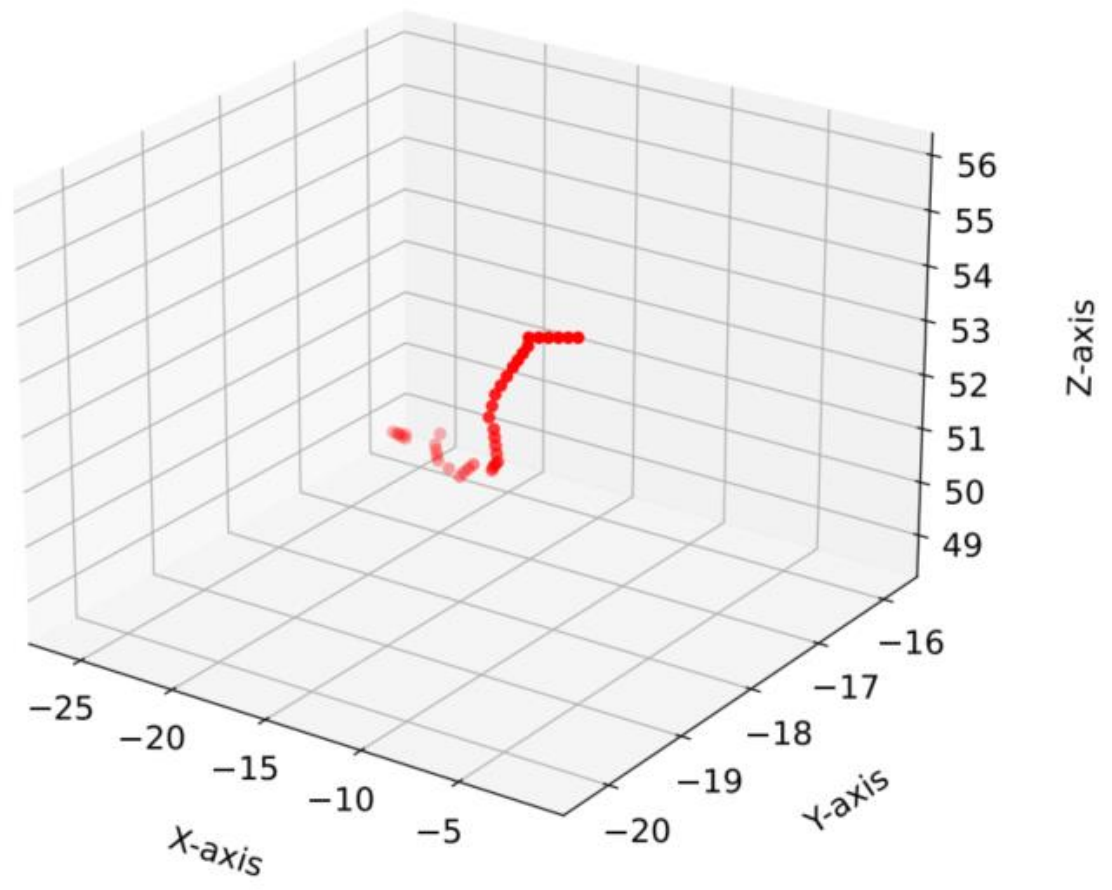

Figure 2: Constructed 3D point cloud of bird's trajectory

Videos from the two cameras resulted in nearly 4 hours of total video footage each. Within those 4 hours of footage, there were events of 1-second duration. After clipping of said events, the dataset came down to 355 clips consisting of birds' flight motion. These motions consist of a bird flying from one perch to another, a bird flying from a hand-held perch to another perch, a bird flying in frame and landing on a perch.

The annotations were done manually using Computer Vision Annotation Tool (CVAT). Annotations were done of the bird itself, head, tail, left-wing, and right-wing. Annotating the whole dataset resulted in 21197 frames with 105985 annotations. Example frames of annotations from both cameras are shown in figure. 3. The annotation's file format is JSON, and the format of the annotation is Microsoft COCO. It should be noted that only the bird in motion was annotated and not the stationary birds. The 3D coordinates for the bird's trajectories are in .mat file format.

In this dataset data could be classified into three categories and sub-category. Category 1: Free flight, leave the bird to decide for itself and fly to another perch from their sitting perch. Few times we touch the hanged parch or kept hand beside the hung perch. They fly through left perch to right perch or right perch to left. Category 2: Place the bird on a hand-held perch and move the perch same direction of the flight or opposite direction of flight or stationery. There are also some random hand perch movement clips. Category 3: Only landing clips where birds are randomly flied and landed on the perch.

In the dataset we investigate various flight motions of four different birds. This will help us to analyze bird flight motions in different criteria. For distinguished between them let's see some identity information and suggest names for four birds in table 1

| No. | Gender | Name | Identity info |
|---|---|---|---|
| 1 | Male | Nigel | Green and older |
| 2 | Male | Hedwig | Small various color of feather |
| 3 | Female | Ava | Yellow with Greenish chest |
| 4 | Female | Joe | Yellow, black marks on solder |

Table 1: Identification of four budgerigars

Table 2, 3, 4 is category wise clip information. Behavior in different situations are also remarked. Table V is list of randomly moved clips, additional information is marked and details have given in note section.

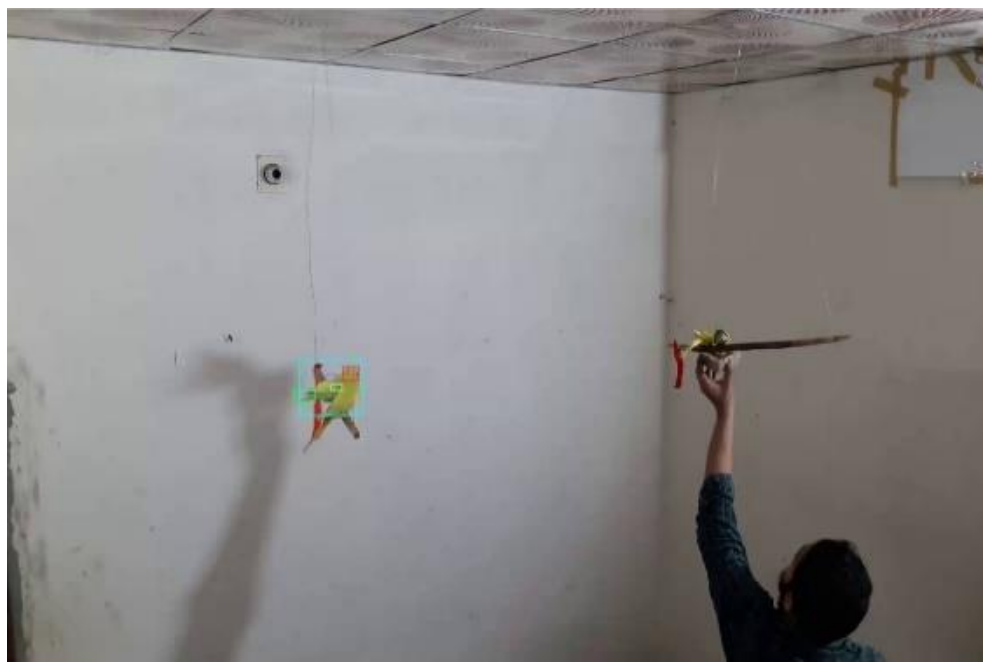
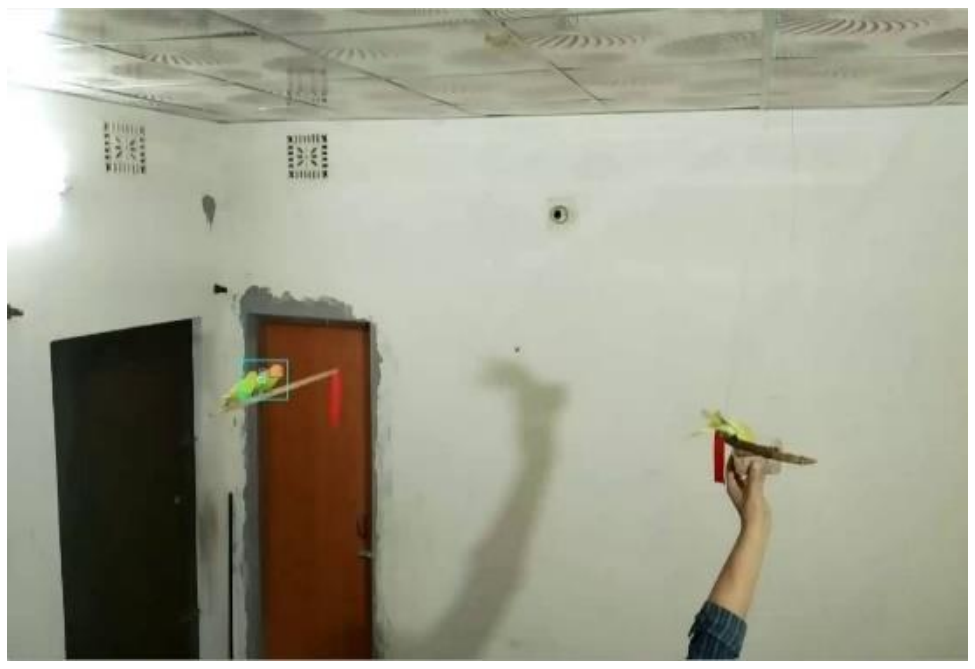

(a) (b)

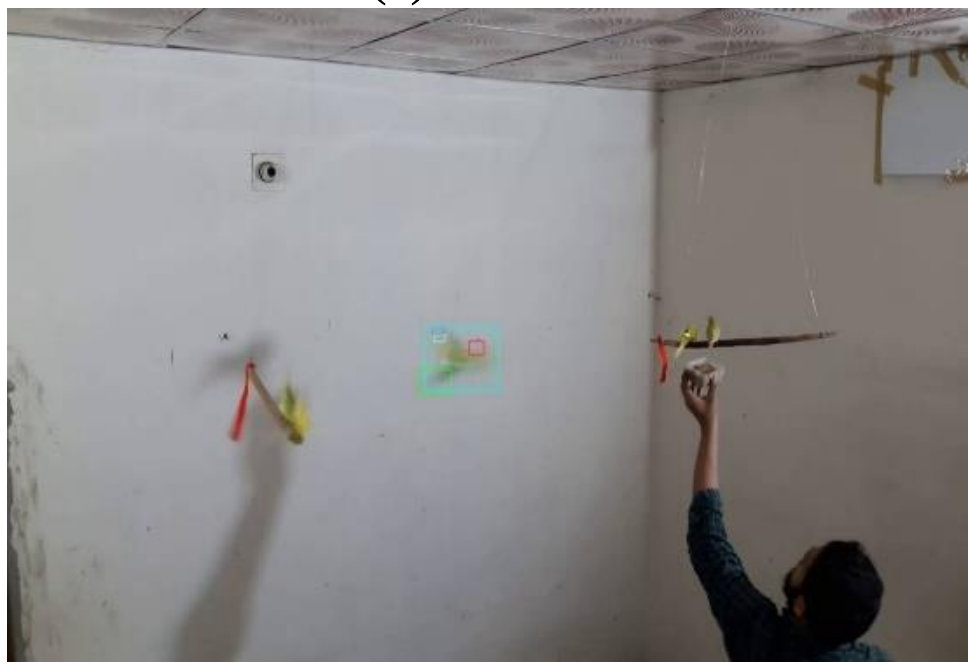
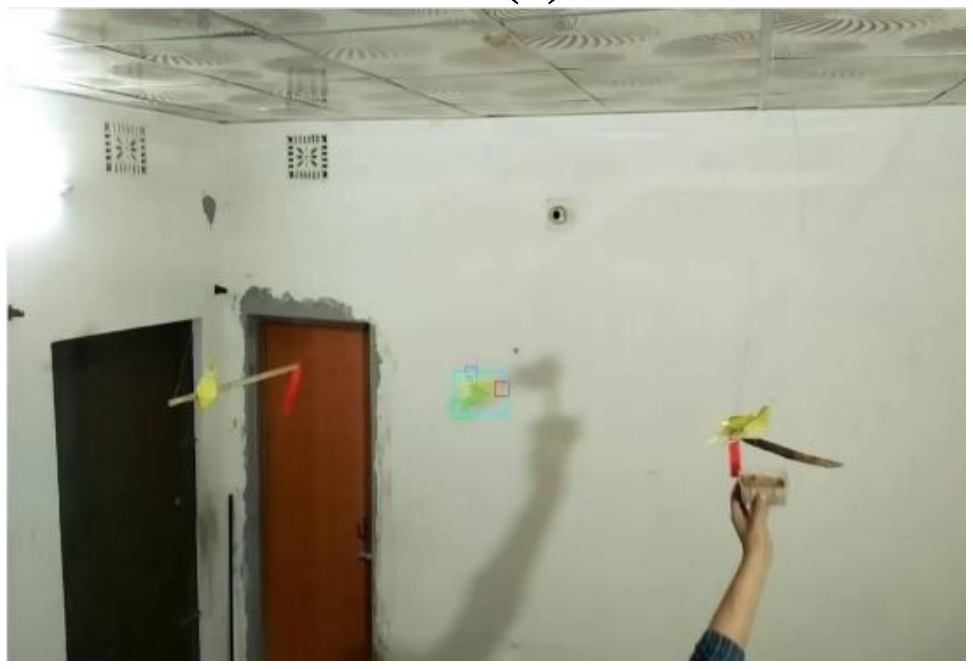

(c) (d)

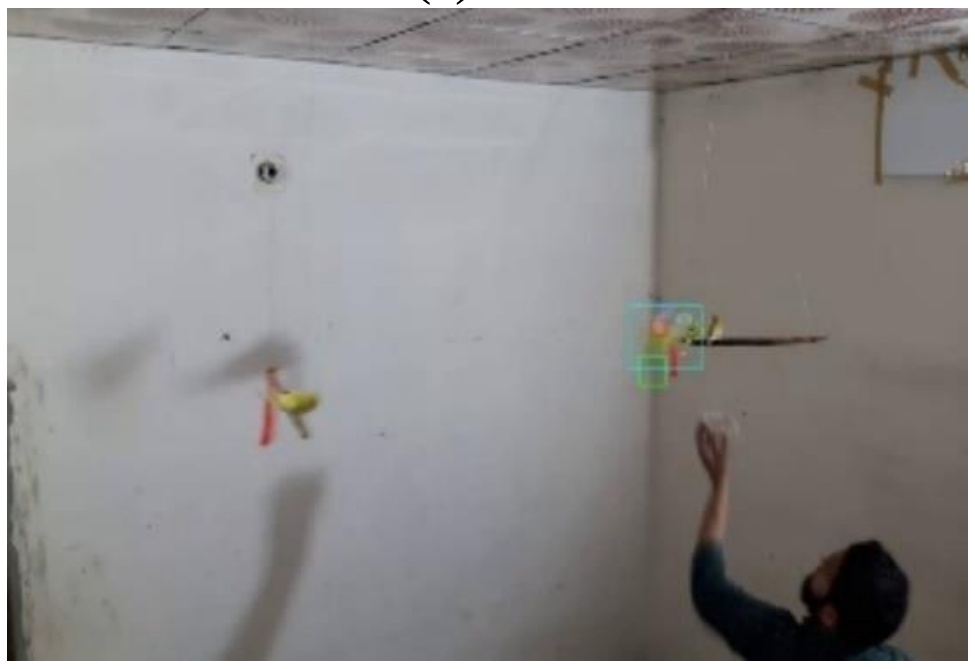
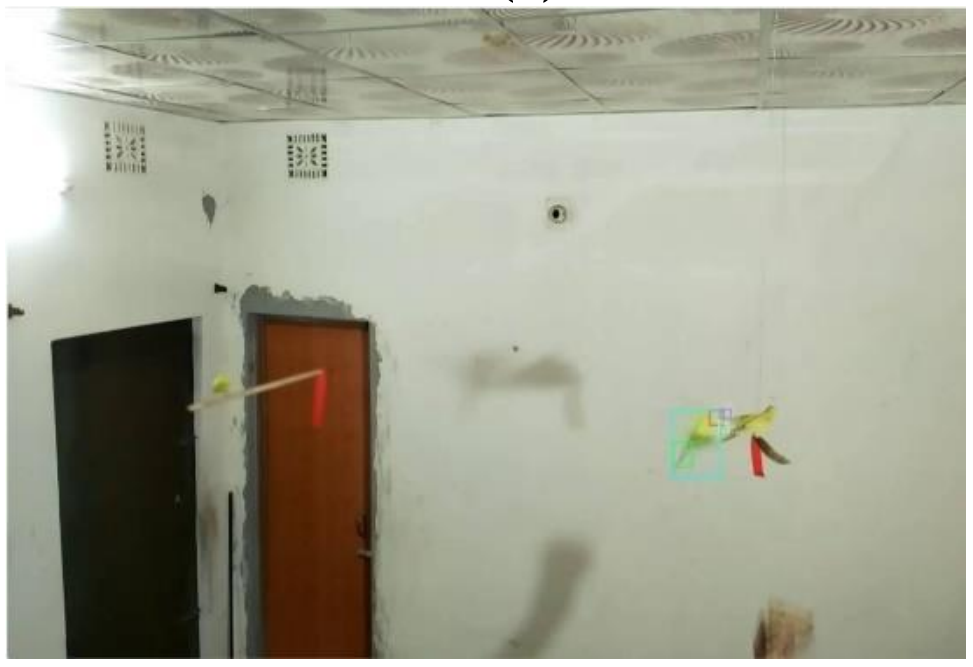

(e) (f)

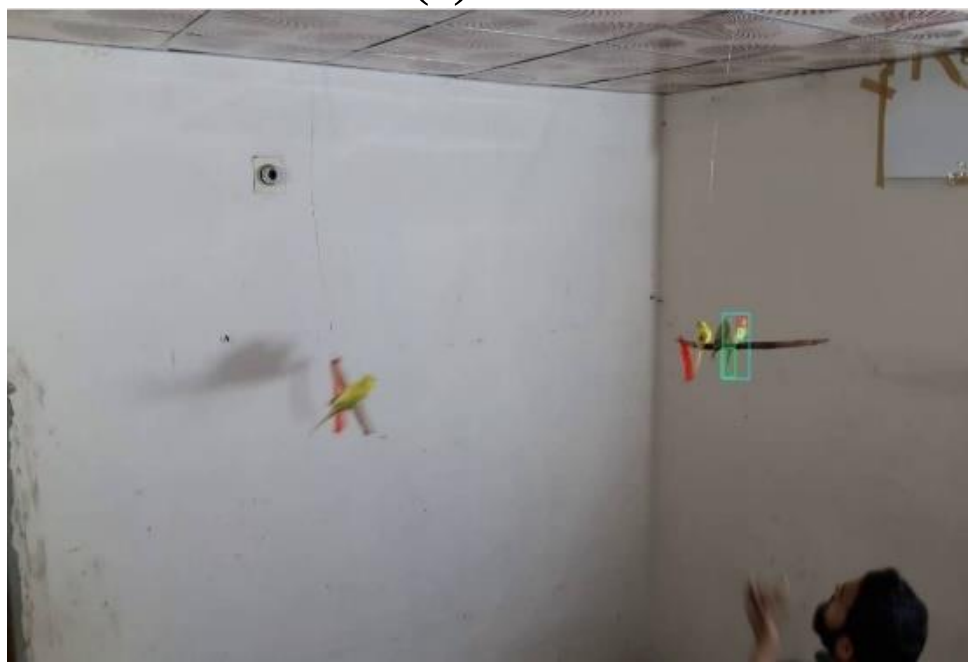
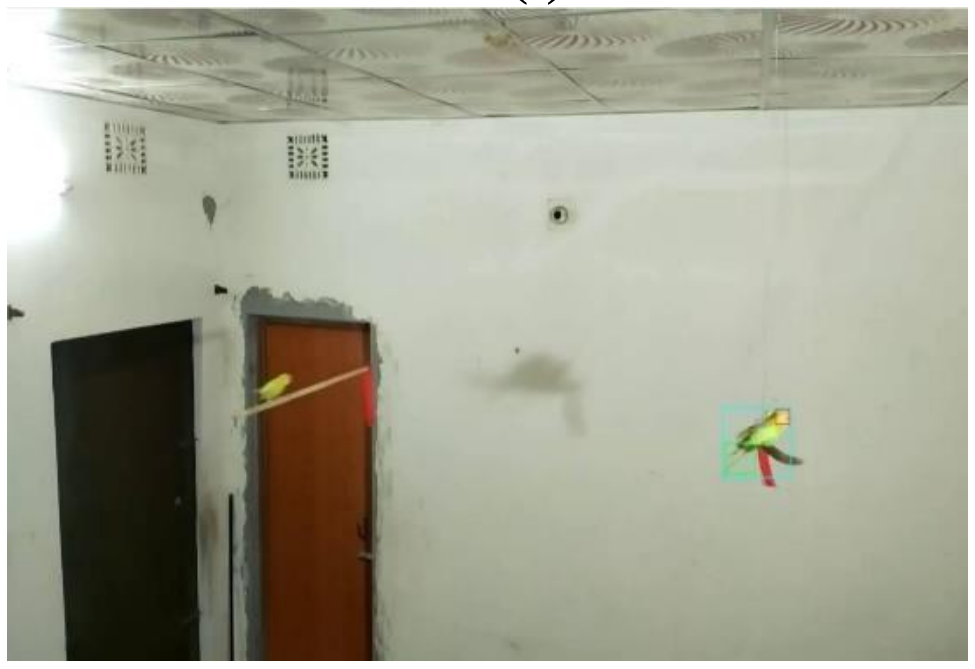

(g) (h)

Figure. 3. Example frames of annotations from both camera 1 (a, c, e, g) and camera 2 (b, d, f, h) of the same clip

| Name | Clip no. Left perch to right perch | Clip no. Right perch to left perch | Clip no. perch |
|---|---|---|---|
| **Nigel** | 11, 12, 17, 20, 23, 26, 29, 33, 34, 40, 41, 68, 71,74, 81, 283, 288, 327,330, 340, 342 | 10, 14, 18, 22, 27, 31, 45, 49, 55, 57, 66, 75, 80, 83, 89, 180, 220(1) 264, 266, 285, 290, 303, 335 | 267(same), 47(hand held), 220(hand held) |
| **Hedwig** | 8, 16, 21, 24, 28, 32, 38, 287, 326, 328, 338, 355 | 329 | 63(same) |
| **Ava** | 2,6, 15, 25, 30, 42, 48, 67 | | |
| **Joe** | 3,4 | | |

**Table 2: Category 1**

| Name | Clip no. perch moved same direction of flight | Clip no. perch moved opposite direction of flight | Clip no. Stationary |
|---|---|---|---|
| **Nigel** | 65rl,181,219,239,277,297,309,321, 324,325rl,337,339,78,98,131, 269,85, 62 | 64rl,166,178,215,230,236,237,247,295,316,318,132,137,159,171, | 194, 199,203, 294, 60, 333, 53, 221,231,235,241(6),291,296,298,301,306,307,310,311,317,319,322rl,72,105,108,133,134,135, 136,138,148,175 |
| **Hedwig** | 51rl,52,169(3),234rl,332,170,205, 244r,245,246,300,304,331rl,35,86, 101,102,114,116,119,158 | 59,249(4),279,282,289,292,293,299 313 314,320,341,347,348,349,115,118,120,149,272 | 278,280,302rl,323,336,96rl,102,103,106,109,111,112,113,151, 163,164, 77 |
| **Ava** | 43rl,50rl,61rl,177,182,206, 265rl,36rl,76rl,100,104,147,160 | 69,70,167,191,196,201,202,209,214,229,243.281,284,139, 144rl,150,274, 275, | 46, 13, 54loop |
| **Joe** | 37rl,39,228,5,7,9rl,19,73,79,87,97rl,121,124,125,126 | 184,190,204,207,208,232,123,130,146rl,161 | 195,200,213 ,129 |

**Table 3: Category 2**

| Name | Only landing clip |
|---|---|
| **Nigel** | 211,217,224rr,227rr,261,262,276,308,92r,143l,142l,153r,173r,270r, (1), 193(2), 1,157 |
| **Hedwig** | 198,212,218,226rr,233rl, 240,248, 250,251,252,253,254,255, 256, 258,259,260, 286(5),315rr,334rl,351lr,352rl,353lr,354lr,82r,90r, 93 94 95,99r,107r,141l,156r,174r,271r,273r,176 |
| **Ava** | 44rl,210 216,223rr,351rr,127r,140l(5) ,143l,172rr, |
| **Joe** | 168, 222rl, 225rr,152r,154rl, 128rl |

**Table 4: Category 3**

| Name | Random |
|---|---|
| Hedwig | Perch<br>-- moved down: 257rl,343<br>--Up and down 84<br>--Move 165<br>--flapping and fly 189,192<br>--back and forward 179,183,346,117 |
| Ava | Perch -- moved down: 238, 155<br>--Forward and back 88 |

Table 5: Randomly hand held perch moved clips

**Notes on Category table:**

(1) landing on hand held perch
(2) flapping and fly
(3) move direction was to right perch, bird moved around right perch avoid obstacle and land in left perch
(4) landed same direction of flight rr
(5) avoid moving perch
(6) make a round of one bird and land

lr = left to right perch
rl = right to left perch
rr = right to right perch
rl = left to right perch
r = right, l = left

From the table II, III, IV, V as we can see that for category 1 Nigel have total 50 clips, Hedwig have total 14 clips, Ava has 10 and Joe have total 2 clips. For category 2 Nigel has 60 clips, Hedwig has 57 clips, Ava has 34 clips, Joe has 29 clips. In category three we observed that Nigel has 17 clips, Hedwig has 37 clips, Ava has 9 clips, Joe has 6 clips. Also, there are some random clips where handheld perch have been moved randomly. From randomly moved clips Hedwig has 10 clips, Ava has 3 clips.

Budges355 is a new annotated dataset is present with 3D trajectory motion of birds. The dataset consists of video clips of birds' motions, with their corresponding frames and annotations. To the best of our knowledge, this is the first dataset that consists of the controlled flight of birds with annotated images tracking their motion and 3D trajectories.

With such a unique dataset, the applications are vast. In this paper we analyzed takeoff, flying and lading motions.

### 3.2. Methods:

As in the dataset there are 3 main categories. For each category we have calculated velocity and the acceleration in three different motions takeoff, flying and landing.

#### 3.2.1. Category 1 and Category 2:

Our motive is to find the velocity and acceleration for three motions takeoff, flying and landing. For calculation we use the formula of kinematics. For acceleration, $a = \frac{2(d-ut)}{t^2}$ have been used where, d = distance between two 3d points = $\sqrt{(x_1 - x_2)^2 + (y_1 - y_2)^2 + (z_1 - z_2)^2}$, u = initial velocity , t = time to travel one point to another, after calculating the acceleration for velocity, ***v = u + at*** have been applied to find the results. Then the average of the velocity and acceleration have calculated for get an overall idea.

In our dataset bird's motion was annotated as four motions sitting, takeoff, flying and landing. For the calculation purpose we finetuned the dataset. We reconstructed the 3D points. We took only the takeoff, flying and landing 3D points based on manual annotation. In figure 4, 3D plots are shown for clip no 330 (Nigel). This clip has been chosen randomly for demonstration purposes.

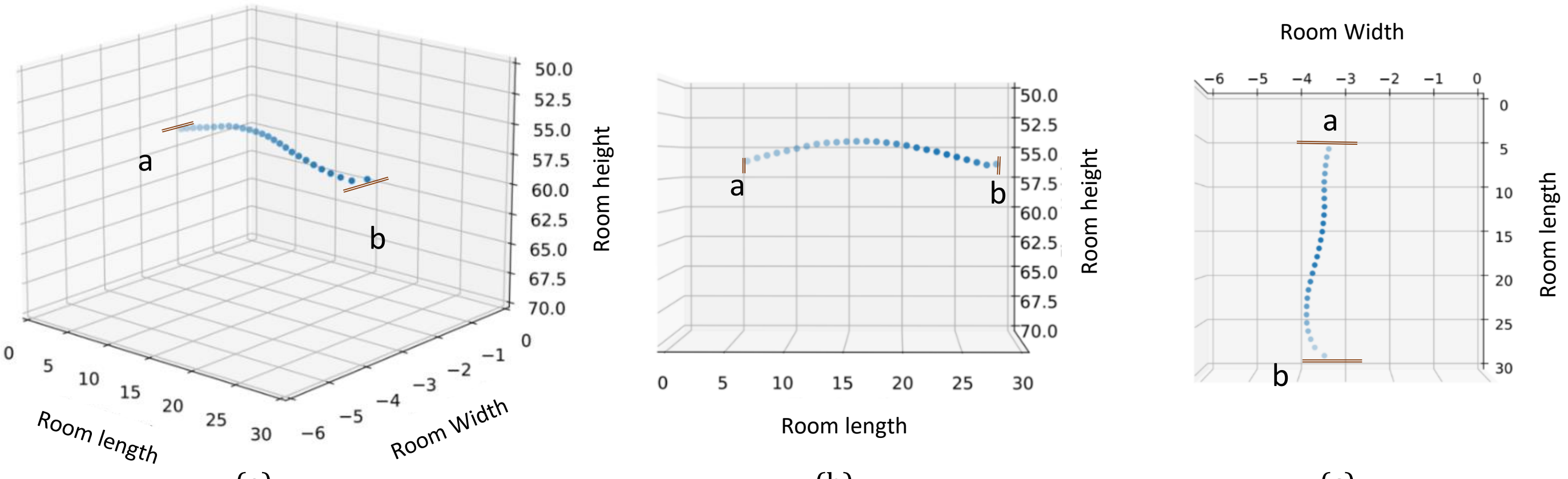

(a) (b) (c)

Figure 4. Finetuned 3D point cloud of bird's trajectory(a), Side view(b), Top view(c)

From the 3D data we separated points for individual motion. For example, in clip 330 takeoff points are shown in figure 5(a), Flying points in 5(b) and landing points in 5(c). All three motions are shown along x (room length) and y (room width) axis. In figure 5(a) takeoff points start from $p_1$ and $p_6$ is the point before the bird starts cruise flying. After point $p_5$ 3d points of flying motions are labeled in serial figure 5(b). From point $p_7$ to point $p_{18}$ was annotated as flying motions point. From $p_{19}$ to point $p_{27}$

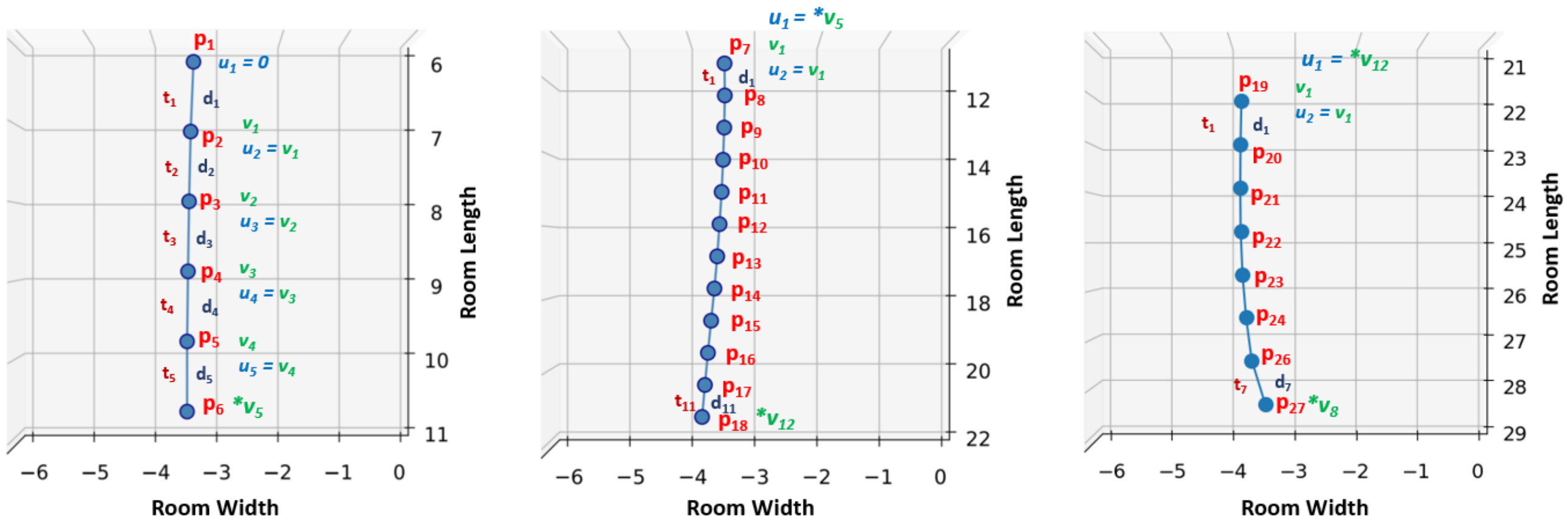

(a) (b) (c)

Figure 5. Takeoff points (a) Flying Points (b) Lading points(c) Top view.

is labeled as landing points.

In figure 5(a) we can see the bird takeoff from point $p1$. Takeoff points start from the position where bird take initiative to takeoff, but did not takeoff yet, In this situation position of the bird is static. So, initial velocity, u = 0, in t time the bird flies through p1 point to p2 in d1 distance. Here the equation of acceleration, $\boldsymbol{a} = \frac{\mathbf{2d1}}{\boldsymbol{t^2}}$ [where u1 = 0] after finding the acceleration we found the velocity v1= at [where, u = 0]. For the next point p2 to point p3 the initial velocity, u2 will be the velocity of point p1 to p2. so, u2 = v1, Equations of the acceleration $\boldsymbol{a} = \frac{\mathbf{2(d2-u2t)}}{\boldsymbol{t^2}}$ and velocity, v2 = u2 +at. Similarly, velocity at point p6 is v5 which is the initial velocity, u5 of next motion flying.

In figure 5(b) flying points are shown. Where after takeoff, point p7 the bird start flying towards the perch for landing. In point p7 initial velocity u1 is the last velocity of takeoff 3D points. so, initial velocity, u1 = ∗v5 which is velocity of takeoff points p5 to p6 in figure 5(a) For flying points start from point p7 to p18. Here for each point to next point acceleration is measured as $a = \frac{2(d-ut)}{t^2}$ where distance denote as $d$ and velocity, v1 = $u1$ +$a1t$. These equations have been used from point $p7$ to point $p18$ to calculate velocity and acceleration. After p18 point the birds starts landing.

Landing motions are annotated when the bird takes the initiative to land like when they use their legs and feet both as air brakes and after landed on perch they close their wings. In figure 5(c) landing points are shown. In figure 5(c) where the bird spread their leg to prepare for landing in point $p19$ and successfully landed in point $p_{27}$. for $p_1$ the initial velocity is $u_1$ [where $u_1$ = $v_{16}$] which is the velocity of flying motions at point $p_{18}$, velocity of landing points is denote as $v1$.

After landing on point $p_{27}$ in In figure 5(c) here the velocity is denote as $V_8$. When the bird landed on point $p_{27}$ with velocity, $V_8$ the hanged perch start oscillating so here we could not say that velocity after landing is 0.

In a particular clip, we calculate the velocity and acceleration from one point to the next point. After calculation, finding from all the points we make an average to for the clip to estimate how they perform on that situation.

### 3.2.2. Category 3

In this category birds were flying randomly and landed on hanged perch. For example, clip 211, constructed in 3d space, in figure 6 shows 3d points where bird randomly flies and landed on the handed perch. For this situation we could not initialize the initial velocity though it is flying we could not say the initial velocity is zero. For this type of clip, we calculated two types of speed of the birds, flying speed and landing speed. First of all we calculated the distance, d, from one point to next and sum up all the points to get the total distance, D (figure 7). after calculated the distance the distance is divided by the total time, t from first point to last. the equation is speed = D/t.

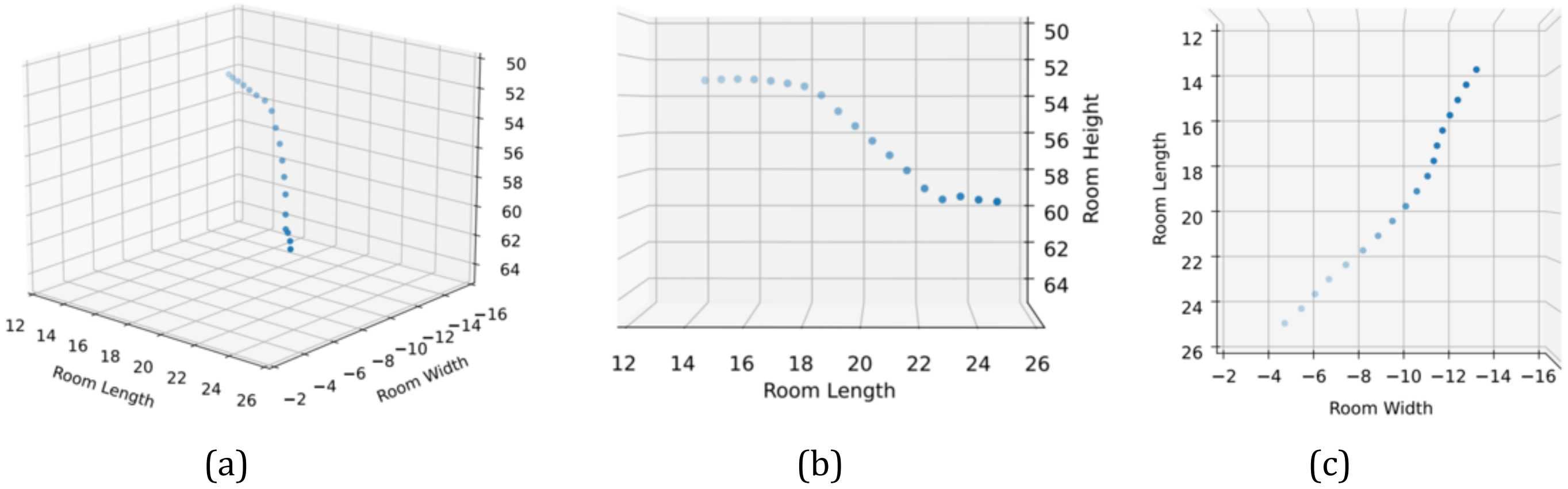

Figure 6. Flying and landing 3D points cloud of bird's trajectory(a), Side view(b), Top view(c)

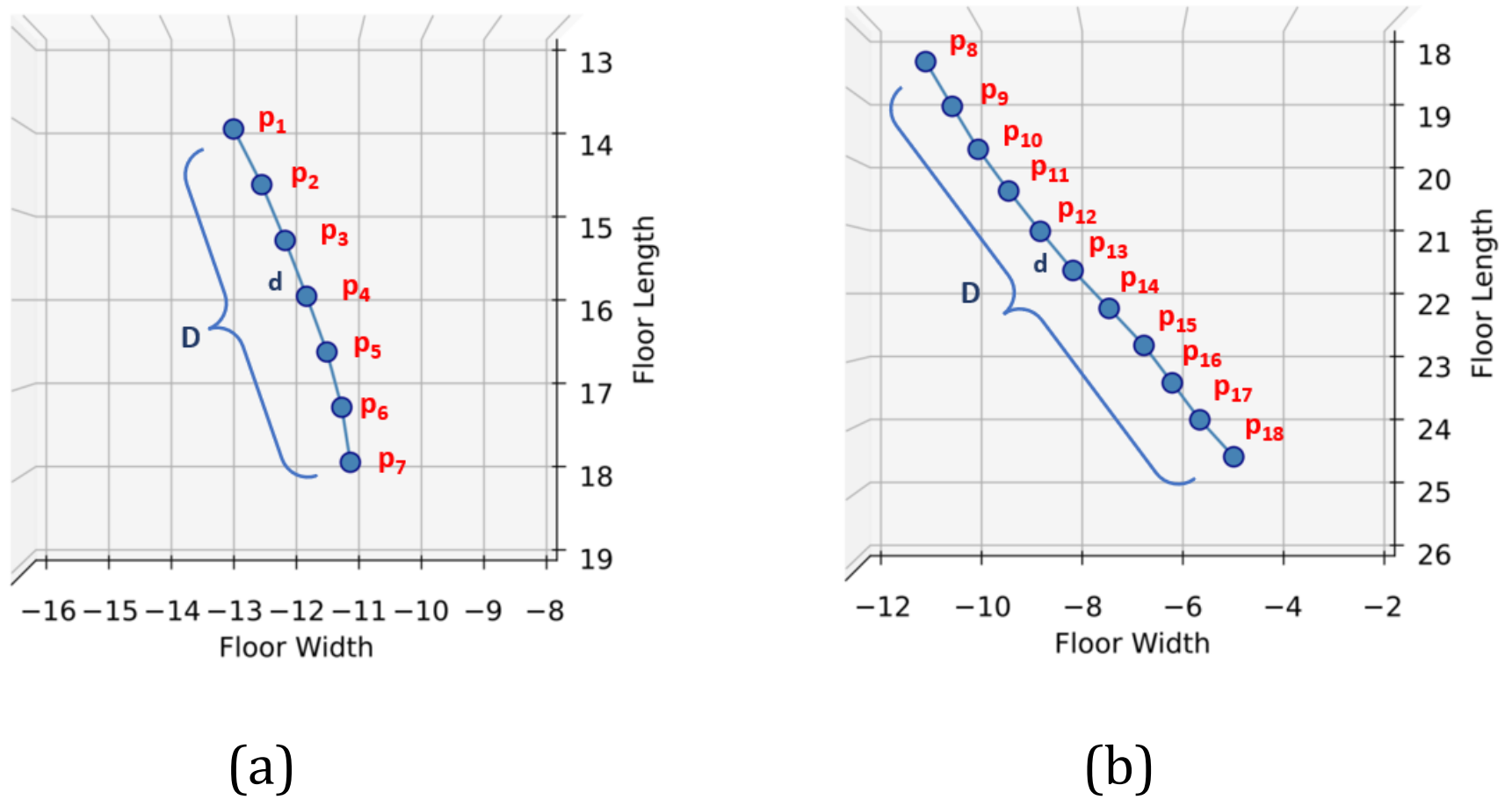

Figure 7. Category 2 Flying points (a), landing points (b) Top view

## 4. Result and Findings

In the methodology section, our investigation is structured around three main categories, each with distinct subcategories, to comprehensively analyze how all four birds, manipulate their velocity and acceleration. We will gradually show the results in this section. Here, we consider the categories as category 1 (velocity & acceleration), category 2 SDF (where hand perch move same direction as flight) velocity and acceleration, category 2 ODF (where hand perch move opposite direction of flight) velocity and acceleration, category 2 S (where hand perch was stationary), Category 3 focusing solely on flying and landing speeds.

Upon calculating the velocity and acceleration from one point to another we calculate the average metrics such as takeoff velocity, takeoff acceleration, flying velocity, flying acceleration, landing velocity, and landing acceleration for each bird, tailored to their unique perspectives. Mean velocity accompanied by standard deviation (SD) of each bird for different categories is demonstrated in figure 7. Additionally, overall mean velocity calculated for each category and shown as black dashed line with SD.

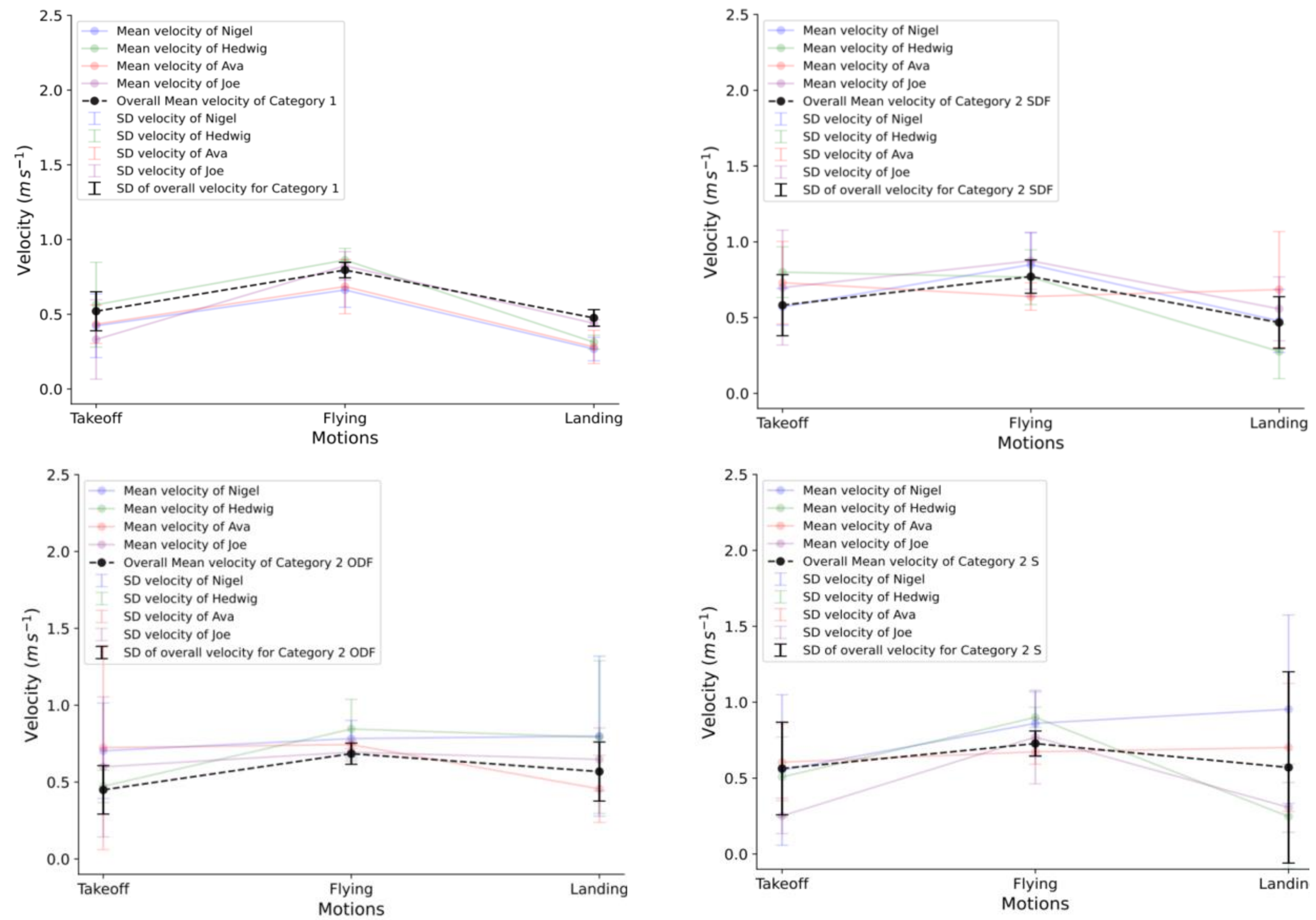

Figure 8: velocity, motions graph for category 1, category 2 SDF, category 2 ODF, category 2 S

Mean acceleration with standard deviation (SD) of each bird for different category is demonstrated in figure 8 also Overall mean acceleration calculated for each category are represented by black dashed lines with SD.

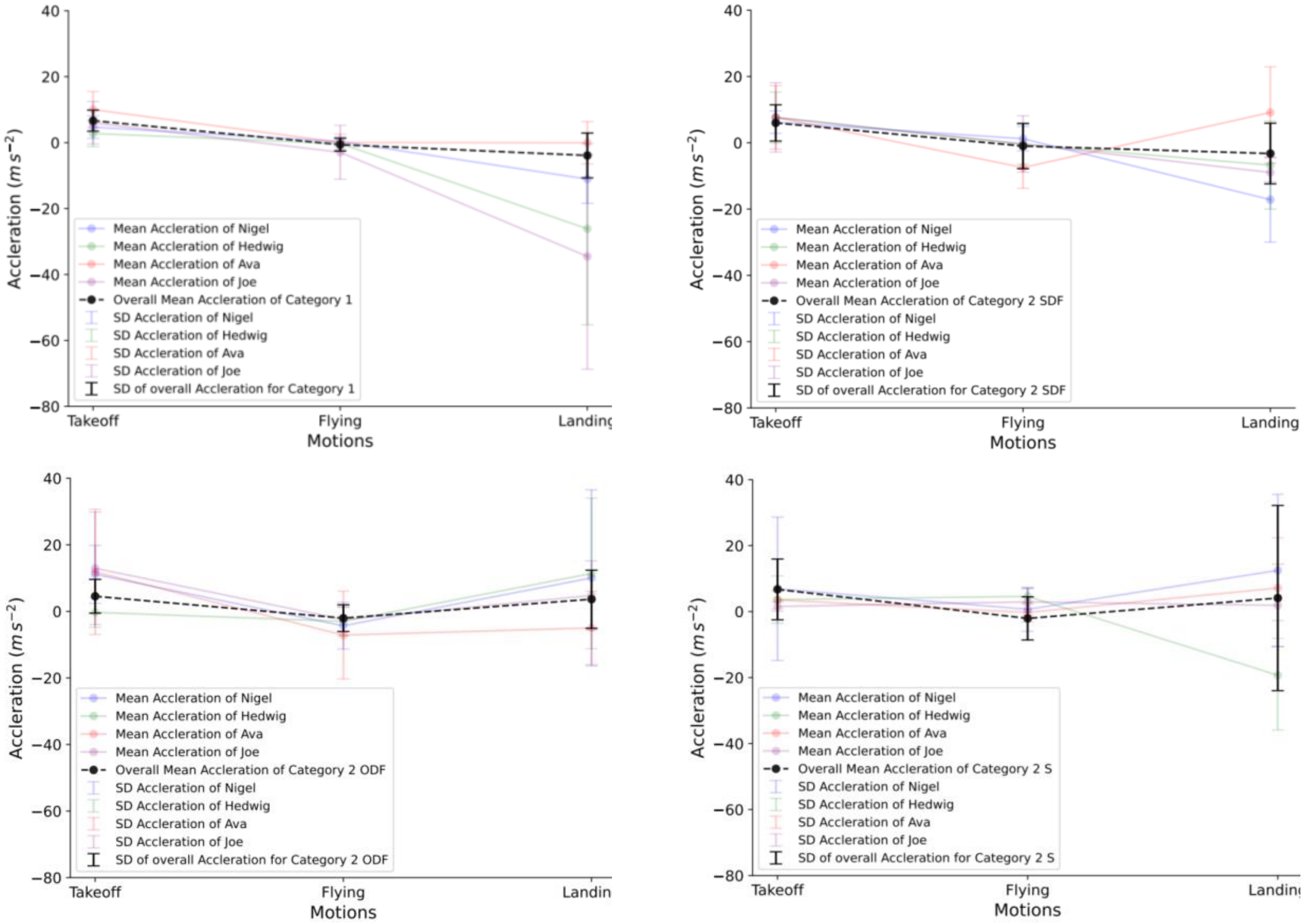

Figure 9: Acceleration, motions graph for category 1, category 2 SDF, category 2 ODF, category 2 S

Category 3, where birds exhibit random flight and land on a hung perch, speed was calculated. The mean speed of each bird with SD is plotted along with Overall mean Speed black dashed line in figure 10.

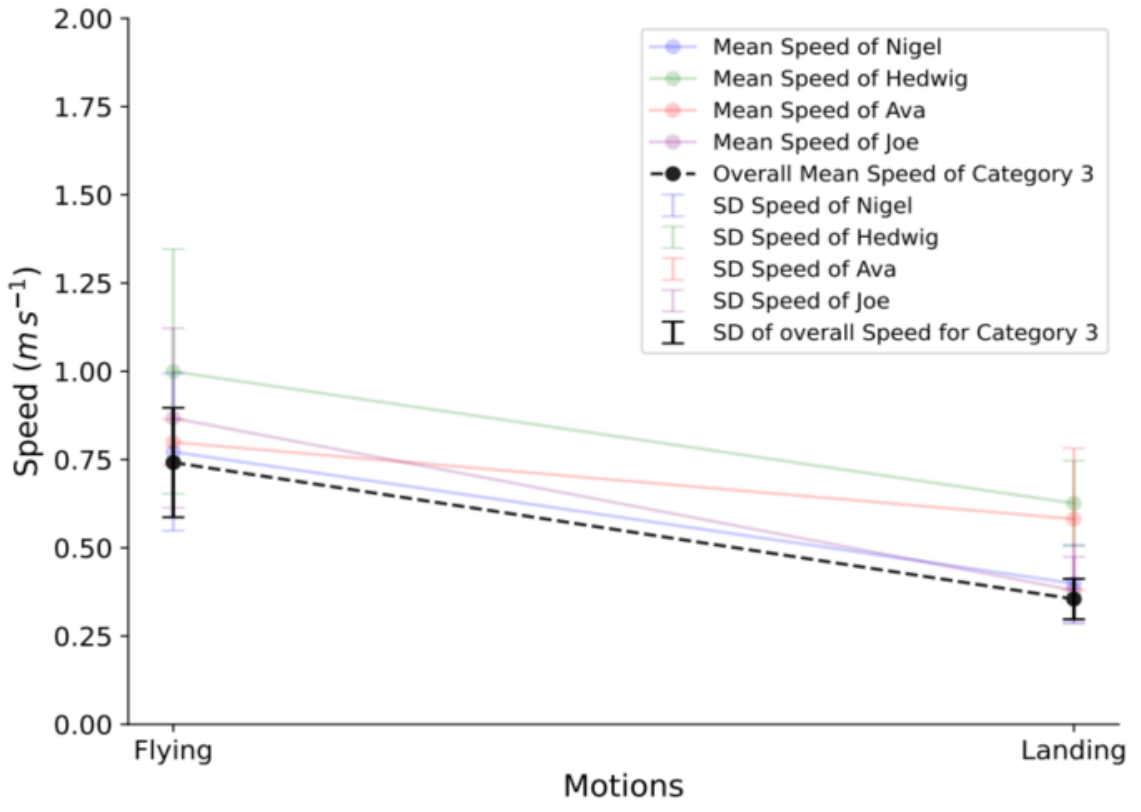

Figure 10: Speed, motions graph for category 3

| | Takeoff Mean | | | | Flyng Mean | | | | Landing Mean | | | |
|---|---|---|---|---|---|---|---|---|---|---|---|---|
| Category Name | Velocity ($ms^{-1}$) | V SD | Accleration ($ms^{-2}$) | A SD | Velocity ($ms^{-1}$) | V SD | Accleration ($ms^{-2}$) | A SD | Velocity ($ms^{-1}$) | V SD | Accleration ($ms^{-2}$) | A SD |
| Category 1 Nigel | 0.42297412 | 0.21324795 | 4.70791441 | 3.42439362 | 0.66254719 | 0.11558673 | 0.32125948 | 1.35505656 | 0.21324795 | 0.0785808 | 11.04461298 | 7.38283818 |
| Category 1 Hedwig | 0.56394039 | 0.28403207 | 2.77253654 | 3.96886547 | 0.86250934 | 0.07833839 | -0.25379799 | 1.7598624 | 0.31336472 | 0.0468853 | -26.13636798 | 29.1283726 |
| Category 1 Ava | 0.43326406 | 0.12852914 | 10.03630138 | 5.47036543 | 0.68555841 | 0.18167065 | 0.05431164 | 2.54870048 | 0.28146242 | 0.11108016 | -0.02138518 | 6.41912383 |
| Category 1 Joe | 0.33126762 | 0.26580221 | 5.99917692 | 6.45426507 | 0.82419344 | 0.09477814 | -2.9227503 | 8.18741842 | 0.43918856 | 0.02416906 | -34.51735131 | 34.2114652 |
| **Overall Mean** | **0.52032221** | **0.13053285** | **6.66062125** | **3.19101692** | **0.79601835** | **0.05155148** | **-0.6125702** | **1.92340722** | **0.47583802** | **0.05545315** | **-3.88534381** | **6.80905597** |
| Category 2 SDF Nigel | 0.57117572 | 0.12131348 | 6.23040138 | 3.39058883 | 0.84935903 | 0.21026271 | 1.23015639 | 3.68957106 | 0.47794666 | 0.20803314 | -17.15586077 | 12.8192897 |
| Category 2 SDF Hedwig | 0.80000904 | 0.16716072 | 7.62080678 | 7.71035906 | 0.76654752 | 0.18056073 | -0.43867521 | 5.9725019 | 0.27645653 | 0.17948355 | -6.74074458 | 13.262205 |
| Category 2 SDF Ava | 0.73025895 | 0.27257719 | 7.66628035 | 9.57214669 | 0.63891797 | 0.09047252 | -7.46166905 | 6.27162065 | 0.68480925 | 0.38271674 | 9.16347503 | 13.7589798 |
| Category 2 SDF Joe | 0.69823787 | 0.37928167 | 7.67353561 | 10.4613292 | 0.87459384 | 0.18858872 | -0.34903797 | 8.50704194 | 0.557823 | 0.21128422 | -8.94916475 | 2.80449436 |
| **Overall Mean** | **0.58192053** | **0.20179754** | **6.00289665** | **5.4679777** | **0.77072786** | **0.10993457** | **-0.97637379** | **6.83423373** | **0.46731927** | **0.17017707** | **-3.23301287** | **9.15090644** |
| Category 2 ODF Nigel | 0.70345354 | 0.31022864 | 11.13687725 | 8.68670106 | 0.78315029 | 0.11784185 | -4.33971357 | 7.00882963 | 0.79869288 | 0.52063275 | 10.05704746 | 26.457408 |
| Category 2 ODF Hedwig | 0.47253233 | 0.10488763 | -0.29880224 | 4.48599106 | 0.84524826 | 0.1928135 | -3.00587926 | 4.33580144 | 0.79237954 | 0.49637823 | 11.38363168 | 22.6199922 |
| Category 2 ODF Ava | 0.72427054 | 0.6638487 | 11.86789556 | 18.8379162 | 0.74386838 | 0.02675165 | -7.13334586 | 13.189429 | 0.45574132 | 0.21826255 | -4.97450241 | 10.9973354 |
| Category 2 ODF Joe | 0.59900717 | 0.45552303 | 12.94903886 | 16.9128997 | 0.69458426 | 0.05711593 | -2.43235165 | 3.66241966 | 0.64773632 | 0.20460215 | 4.79422231 | 10.3804333 |
| **Overall Mean** | **0.44945449** | **0.15775771** | **4.53882639** | **5.06676212** | **0.68471137** | **0.06888283** | **-2.03231669** | **4.00698375** | **0.56839832** | **0.19207938** | **3.66068141** | **8.73663787** |
| Category 2 S Nigel | 0.55380924 | 0.49583552 | 6.9127625 | 21.7365736 | 0.86061042 | 0.20727121 | 0.69602155 | 6.65765979 | 0.95388052 | 0.62062014 | 12.48458417 | 23.065958 |
| Category 2 S Hedwig | 0.50768761 | 0.26286158 | 3.63952344 | 7.19943327 | 0.90038394 | 0.06660173 | 4.62340549 | 2.35986096 | 0.24725014 | 0.34364255 | -19.29834601 | 16.567907 |
| Category 2 S Ava | 0.60487616 | 0.25342021 | 3.51797973 | 2.6740513 | 0.67378046 | 0.08138384 | -0.26354929 | 2.90192782 | 0.70216566 | 0.42181717 | 7.14383082 | 15.2506391 |
| Category 2 S Joe | 0.25117061 | 0.11686904 | 1.5554048 | 1.42671736 | 0.7707164 | 0.30870659 | 2.94214988 | 4.10840617 | 0.30738204 | 0.1638952 | 1.89652634 | 12.4825965 |
| **Overall Mean** | **0.56342392** | **0.30505699** | **6.7340028** | **9.20320059** | **0.72764569** | **0.08276069** | **-2.05960866** | **6.55592274** | **0.57043021** | **0.62984302** | **4.11135146** | **28.0643469** |

Table 6: Takeoff, Flying, landing motion mean velocity, acceleration, Standard Deviation (SD) of mean velocity (V SD) and mean acceleration (A SD). For each category overall mean of velocity, acceleration, and Standard Deviation (SD)

For all categories data tables are added in appendix section with proper table headings. The tables tell us all the individual average motions velocity and acceleration values of each clip and the graphs accumulate all those values in order to summarize the table and give us an idea about the trend the birds are following. The graphs show a consistent trend amongst the clips. There can be seen a repetitive pattern in all the graphs for all categories. This consistency of the graphs is very reassuring.

According to theory a bird will accelerate, or its velocity will increase while it's taking off and it will decelerate, or its velocity will decrease when it is landing. Looking at figure 8,9 mean velocity of takeoff motion is increased; in flying motion they try to consist of their speed in a range in time of landing the velocity is decreased, accordingly deceleration occurred. The practical data for the theory can be seen in the play. For category 3 figure 10 where flying speed higher and when it's time to land it speed down for landing is clearly seen.

Our analysis aims to enhance the dataset so that it has more ground truth. To enhance the dataset, we needed to extract data out of the dataset and have a consistent result in our extraction for it to be labelled as ground truth.

## 5. Discussion:

This paper presents a comprehensive analysis of the dataset, a pioneering collection capturing the controlled flight of Budgerigars with annotated images tracking their motion. The dataset comprises 355 clips, showcasing various bird movements and speed variations across different scenarios. Our primary objective was to extract and analyze velocity and acceleration data from the four birds' takeoff, flying, and landing motions, aiming to enhance the dataset with reliable ground truth for potential applications in neural network training.

Our analysis revealed intriguing insights into the kinematics of bird flight. As anticipated, the findings generally align with the theoretical expectations of bird behavior during flight. During takeoff, the birds exhibited an increase in velocity, followed by a more controlled and consistent speed during flying, and a subsequent decrease in velocity upon landing. This observed pattern is consistent with existing theoretical models of bird flight, providing validation for the dataset's accuracy.

Comparisons with previous studies in avian biomechanics and flight dynamics further support the robustness of our findings. Research on birds' optical flow-based navigation in narrow tunnels, their velocity adjustments in varying tunnel widths, and collision avoidance behaviors during head-on encounters resonates with our observations. Additionally, studies investigating bird landing dynamics based on surface properties and the use of visual edges for targeting align with our exploration of Budgerigar flight.

However, it is crucial to note certain anomalous data points that deviate from the expected patterns. These anomalies highlight the complexity and variability inherent in avian flight, underscoring the need for further in-depth investigations to fully comprehend the intricacies of bird behavior.

Looking forward, the enhanced of our work which enriched with our extracted ground truth, opens avenues for advanced research in the realm of bio-inspired computer vision. Future studies could explore applications in unmanned aerial vehicles (UAVs) and autonomous robotics, leveraging the dataset's detailed insights into bird flight dynamics. Comparative analyses with additional avian species and diverse environmental conditions could further broaden the dataset's utility and contribute to a more comprehensive understanding of visually directed flight in flying species.

In conclusion, our study not only validates the fidelity of the dataset but also serves as a foundation for future research endeavors in the dynamic field of biomimicry and avian-inspired technologies. The complexities uncovered in bird flight behaviors beckon further exploration, driving the quest for innovative solutions and advancements in autonomous aerial systems and robotics.

## Appendix:

| Category 1 Nigel | | | | | | |
|---|---|---|---|---|---|---|
| | Velocity ($ms^{-1}$) | | | Accleration ($ms^{-2}$) | | |
| **clip** | **Takeoff** | **Flying** | **Landing** | **Takeoff** | **Flying** | **Landing** |
| 10 | 0.101871431 | 0.59882565 | -0.145219449 | 2.827299243 | 1.370008096 | -17.13347448 |
| 11 | 0.082914868 | 0.617643293 | 0.200447312 | -0.009212468 | -0.519128519 | -10.83335751 |
| 12 | 0.598416842 | 0.729080835 | 0.319487922 | -0.099833275 | 0.052860495 | -14.59705719 |
| 14 | 0.360230084 | 0.710741563 | 0.61395039 | 9.468928086 | -0.380307139 | 2.189639616 |
| 17 | 0.254745204 | 0.704728998 | 0.138752111 | 2.370610376 | -2.770638513 | -1.479305971 |
| 18 | 0.49028971 | 0.94807725 | 0.435857093 | 8.961291283 | 1.481370923 | -7.716546731 |
| 20 | 0.052552506 | 0.945624436 | 0.405356718 | -0.094434642 | 2.501777579 | -3.693218018 |
| 22 | 0.253422728 | 0.733509086 | 1.209725556 | 5.859418991 | -2.098130066 | 33.83038937 |
| 23 | 0.592958236 | 0.747486805 | 0.401411582 | 16.49910999 | -2.838063394 | 0.298779375 |
| 26 | 0.180002629 | 0.857166823 | 0.345448417 | 2.438129228 | 3.569815396 | -4.166599296 |
| 27 | 0.287372867 | 0.777026068 | 0.989529588 | -0.140062921 | 2.688207634 | 11.86281454 |
| 29 | 0.070867747 | 0.893273846 | 0.509018956 | -0.001328198 | 3.329181062 | -2.18509843 |
| 31 | 0.17745618 | 0.726902963 | 0.513752642 | 3.489953892 | 0.528373468 | -11.52816305 |
| 33 | 0.757769683 | 0.928004412 | 0.425192834 | 5.332621086 | 0.231499461 | -1.835425603 |
| 34 | 0.231471563 | 0.803734527 | 0.27786 | 5.274483049 | 1.992180197 | -3.153831076 |
| 40 | 0.562966204 | 1.195248263 | 0.340363546 | -11.21772267 | 18.2170943 | -23.24683284 |
| 41 | 1.132191393 | 0.772993661 | 0.514927219 | 15.69150934 | -0.036441489 | -17.38836281 |
| 45 | 0.810172842 | 0.930170988 | 1.938285731 | 7.412058111 | -13.15058959 | 47.38349716 |
| 47 | 0.617156978 | 0.594661692 | 0.530965388 | 4.405660681 | -1.287815679 | -1.33318567 |
| 49 | 0.710548449 | 0.839075703 | 0.741642612 | 17.91892634 | -3.215641663 | 11.38811973 |
| 55 | 0.638173217 | 0.95761202 | 0.488090448 | 3.261749995 | 1.084308437 | 0.381014988 |
| 57 | 0.906599701 | 0.840506834 | 0.679294201 | 17.49362388 | 0.113156105 | -1.067672627 |
| 66 | 0.746946402 | 1.381637845 | 0.285535363 | -2.491862525 | 17.25310509 | -32.66605057 |
| 68 | 0.568308717 | 0.580474882 | 0.266112042 | 9.791893909 | -9.733247484 | -11.04218887 |
| 71 | 0.843579254 | 0.789425784 | 0.328580872 | 11.55667153 | -4.037398997 | -3.52653853 |
| 74 | 0.369458857 | 0.870961297 | 0.601351235 | -2.548239383 | 5.094425612 | 0.781795156 |
| 75 | 0.486938649 | 0.795688694 | 1.112292079 | 6.303565796 | -0.833453193 | 23.81814482 |
| 80 | 0.054724121 | 0.699453872 | 0.532285841 | 0.006162243 | 0.916786346 | 0.233553203 |
| 81 | 0.612616288 | 0.858392651 | 0.537395583 | 10.23672391 | 0.454496559 | -15.67589845 |
| 83 | 0.478608567 | 0.855831418 | 0.482379332 | 5.223307467 | 1.695499258 | -5.900368609 |
| 89 | 0.612128859 | 0.726197713 | 0.635999812 | 0.891684255 | -5.385169782 | -7.532547047 |
| 220 | 1.058296999 | 0.59974493 | 0.434044469 | 20.42980049 | -6.27220711 | 7.317034897 |
| 220 | 1.058296999 | 0.59974493 | 0.434044469 | 20.42980049 | -6.27220711 | 7.317034897 |
| 220 | 1.058296999 | 0.59974493 | 0.434044469 | 20.42980049 | -6.27220711 | 7.317034897 |
| 220 | 1.058296999 | 0.59974493 | 0.434044469 | 20.42980049 | -6.27220711 | 7.317034897 |
| 264 | 0.642121087 | 1.049315795 | 0.327832181 | 16.545858 | 8.345415628 | -8.969000261 |
| 266 | 0.326852351 | 0.701400404 | 1.214065798 | 5.471655262 | -11.23849168 | 47.21860624 |
| 267 | 0.588086365 | 0.561265162 | -1.389576566 | 11.80400756 | 3.465283033 | -102.531709 |
| 283 | 0.30751452 | 0.665181963 | 0.222123483 | 4.279889458 | -1.220975888 | -4.028289465 |
| 285 | 0.434657642 | 0.894683579 | 0.680502024 | 2.624390358 | 1.949507791 | 5.998828615 |
| 288 | 0.593794351 | 0.975122908 | 0.461699569 | 6.030766325 | -5.026416978 | -7.753502542 |
| 290 | 0.269912738 | 0.559043964 | 0.421752189 | 4.537845745 | -4.123272909 | -2.193391247 |
| 303 | 0.370399873 | 0.848184118 | 0.447577318 | 1.497946626 | -1.774095327 | -5.304776915 |
| 327 | 0.27762827 | 0.647867153 | 0.357509976 | 1.073284924 | 1.423855148 | -1.285018708 |
| 330 | 0.341023319 | 0.791808229 | 0.304043774 | -0.660028394 | -7.085661023 | -1.018264082 |
| 335 | 0.963499149 | 0.713207796 | 0.38697885 | 9.549739762 | -12.09791589 | -2.081301327 |
| 340 | 0.754293321 | 0.767138579 | 0.192483179 | 16.73754485 | -1.211354228 | -10.04243584 |
| 342 | 0.598231953 | 0.733012557 | 0.602613636 | 8.637673111 | -2.150145726 | 8.291363794 |
| **Mean** | **0.528013827** | **0.785757746** | **0.471288672** | **6.790884629** | **-0.823853666** | **-2.499265137** |

| Category 1 Hedwig | | | | | | |
|---|---|---|---|---|---|---|
| | Velocity ($ms^{-1}$) | | | Accleration ($ms^{-2}$) | | |
| **clip** | **Takeoff** | **Flying** | **Landing** | **Takeoff** | **Flying** | **Landing** |
| 8 | 0.437043619 | 0.526339101 | 0.178659172 | 3.153668806 | 0.437371627 | -2.926407971 |
| 16 | 1.005681862 | 0.927002074 | 0.386873555 | 1.164117077 | -3.204005804 | -76.18045808 |
| 21 | 0.552152436 | 0.454093186 | 0.409681711 | 12.96624439 | -0.47191224 | 8.46731595 |
| 24 | 0.252188379 | 0.812622915 | 0.451889059 | 3.160642432 | 3.223970092 | -17.63327066 |
| 28 | 0.11042123 | 0.513186083 | 0.294607644 | -0.002813603 | -0.433568263 | 6.451247394 |
| 32 | 0.105029918 | 0.778133746 | 0.395661375 | 0.442281945 | 0.637850482 | -14.19143764 |
| 38 | 0.934168793 | 1.121680366 | 0.544211494 | -3.760394639 | 6.381172201 | -12.20143858 |
| 63 | 1.491337501 | 0.725778768 | 0.164512755 | 29.31686553 | -0.002461345 | -13.62518358 |
| 287 | 0.325754263 | 0.745324799 | 0.482345526 | -0.433111722 | 3.022687739 | -3.232295148 |
| 326 | 0.137962244 | 0.841721224 | 0.404997581 | -0.019298456 | 0.850568823 | 0.833551101 |
| 328 | 0.458686631 | 1.071812626 | 0.198596145 | 2.411883193 | 7.40974741 | -20.04840586 |
| 329 | 0.347100594 | 1.000936217 | 0.42105436 | 0.236763956 | 3.290556207 | -4.657548924 |
| 338 | 0.478486361 | 0.728445843 | 0.352735777 | 6.403381331 | -0.572727484 | 0.055036695 |
| 355 | 0.254638969 | 0.944973295 | 0.483734053 | 0.316390866 | 3.878484038 | -12.88828271 |
| **Mean** | **0.492189486** | **0.79943216** | **0.369254301** | **3.954044365** | **1.746266677** | **-11.55554129** |

| Category 1 Ava | | | | | | |
|---|---|---|---|---|---|---|
| | Velocity ($ms^{-1}$) | | | Accleration ($ms^{-2}$) | | |
| **clip** | **Takeoff** | **Flying** | **Landing** | **Takeoff** | **Flying** | **Landing** |
| 2 | 0.666361282 | 0.66235243 | 0.353774104 | 6.709170241 | 2.467030133 | -7.548228376 |
| 6 | 0.406394935 | 0.886753189 | 0.264028906 | 9.60266509 | 0.848294408 | -9.620097754 |
| 15 | 0.49289454 | 0.897853036 | 0.295953431 | 14.77204937 | 3.405485674 | -7.052165528 |
| 25 | 0.689061981 | 0.714332957 | 0.460319899 | 14.93132297 | -14.49399858 | -82.22556523 |
| 30 | 0.170458895 | 0.823456035 | 0.502643648 | 4.040208376 | 1.495616311 | -7.116355166 |
| 42 | 0.739272542 | 0.930608062 | 0.8262243 | 10.94618659 | -2.358689749 | 14.99915026 |
| 48 | 0.631798408 | 1.050114288 | 0.218785051 | 5.827275628 | 6.529050563 | -15.09491663 |
| 67 | 0.827744049 | 0.624810077 | 0.24482876 | 16.11209639 | -9.061045652 | -5.2890507 |
| **Mean** | **0.577998329** | **0.823785009** | **0.395819763** | **10.36762183** | **-1.396032111** | **-14.86840364** |

| Category 1 Joe | | | | | | |
|---|---|---|---|---|---|---|
| | Velocity ($ms^{-1}$) | | | Accleration ($ms^{-2}$) | | |
| **clip** | **Takeoff** | **Flying** | **Landing** | **Takeoff** | **Flying** | **Landing** |
| 3 | 0.505576718 | 0.843853922 | 0.337269593 | 8.97803108 | -1.100235309 | -22.87045805 |
| 4 | 0.245267906 | 0.907201259 | 0.283068514 | 0.423197259 | 1.287658929 | -4.147858915 |
| **Mean** | **0.375422312** | **0.875527591** | **0.310169054** | **4.700614169** | **0.09371181** | **-13.50915848** |

| Category 2 SDF Nigel | | | | | | |
|---|---|---|---|---|---|---|
| | Velocity ($ms^{-1}$) | | | Accleration ($ms^{-2}$) | | |
| **clip** | **Takeoff** | **Flying** | **Landing** | **Takeoff** | **Flying** | **Landing** |
| 62 | 0.164840877 | 0.977376065 | 0.658984614 | -5.911979707 | -0.103150832 | 9.783866732 |
| 65 | 0.738972104 | 0.978671432 | 0.21860462 | 9.981749687 | 3.345880711 | -27.69752193 |
| 78 | 0.367846699 | 0.675710176 | 0.30408834 | 0.06429051 | -0.733958541 | -8.423217023 |
| 85 | 0.677028281 | 0.806582992 | 0.782068428 | 6.207757779 | -3.312695256 | 11.07663918 |
| 98 | 0.740894055 | 0.72567019 | 0.206317637 | 8.705387446 | -1.577249249 | -5.423803805 |
| 131 | 0.345617008 | 0.515280664 | 0.782227392 | 3.457897198 | -1.622611945 | 7.523552864 |
| 219 | 0.810112323 | 0.951421564 | 0.402852956 | 6.909490898 | 7.363706649 | -3.004667992 |
| 239 | 0.59724418 | 0.708793507 | 1.211918311 | -3.000297658 | -8.596868893 | 25.70593423 |
| 269 | 0.536507167 | 0.510830395 | 0.440809818 | 9.773418792 | -8.67396892 | -2.35027196 |
| 277 | 0.147878224 | 0.776382882 | 0.293243591 | 2.219304462 | -8.873734305 | -4.855200226 |
| 297 | 0.495348797 | 0.840896876 | 0.324489024 | 12.75542886 | 0.233437798 | -5.236319982 |
| 309 | 0.285893841 | 0.718034999 | 0.41959687 | 2.919539252 | 0.083821572 | 0.848432864 |
| 321 | 0.346494072 | 0.818283158 | 0.520701389 | 2.862864509 | -1.339911334 | 7.853315844 |
| 324 | 0.213953372 | 0.781592899 | 0.22789066 | 3.014894514 | 0.647027824 | -5.868236336 |
| 325 | 0.454467664 | 1.699405552 | -0.401739181 | 1.396228537 | 26.30304794 | -61.20551693 |
| 337 | 0.591017998 | 0.824616906 | 0.247292363 | 6.65558426 | 2.346220703 | 0.446476433 |
| 339 | 0.774728342 | 0.877301064 | 0.302527701 | 0.206914749 | 3.369231627 | -7.16263469 |
| **Mean** | **0.487579118** | **0.834520666** | **0.408345561** | **4.012851417** | **0.521072091** | **-3.999363102** |

| Category 2 SDF Hedwig | | | | | | |
|---|---|---|---|---|---|---|
| | Velocity ($ms^{-1}$) | | | Accleration ($ms^{-2}$) | | |
| **clip** | **Takeoff** | **Flying** | **Landing** | **Takeoff** | **Flying** | **Landing** |
| 35 | 0.420954314 | 0.56895822 | 0.112661397 | 6.70237457 | -0.555670727 | -2.418271026 |
| 51 | 0.400728236 | 0.491370628 | 0.445813504 | 7.218133275 | -2.482378468 | -13.55529479 |
| 52 | 1.069229095 | 0.830834856 | -0.029139288 | 17.61967242 | 1.198519131 | -26.35153179 |
| 86 | 0.161623121 | 0.736065522 | 0.25963255 | 1.896341226 | 2.807245896 | -34.92108332 |
| 101 | 0.758462697 | 0.614682377 | 0.442683189 | 10.22702606 | -2.801558135 | 0.080026526 |
| 102 | 1.12585167 | 0.803389697 | 0.568911407 | 19.41546674 | -4.020267654 | 1.433271888 |
| 114 | 0.756708345 | 0.696608545 | 0.813892388 | 9.892962892 | 0.199313377 | 0.209249545 |
| 116 | 0.307542989 | 0.854677813 | 0.359296388 | 0.03302683 | 1.435747796 | -5.98516908 |
| 119 | 0.362961511 | 0.767364746 | 0.351769672 | 1.785468825 | 2.114286527 | -1.821409246 |
| 158 | 0.36441202 | 0.844945425 | 0.278117348 | 0.19203782 | 4.328320898 | -9.650192976 |
| 169 | 0.561139451 | 0.60935559 | 0.850580846 | 3.951250019 | -1.056913391 | 19.00632474 |
| 170 | 0.330149085 | 0.558792137 | 0.394361748 | -0.66150435 | -0.33329792 | 3.018009935 |
| 205 | 0.642087893 | 0.874168373 | 0.430966439 | 1.681716095 | 3.441703013 | -4.754063739 |
| 234 | 1.035655903 | 0.933514148 | 0.839269539 | 19.40793938 | 0.943805693 | -9.383811535 |
| 244 | 1.185663588 | 0.64336936 | 0.691939779 | 27.49411313 | -13.51570326 | -3.024649805 |
| 245 | 0.867977116 | 0.774639987 | 0.514849778 | 14.95467493 | -7.188445072 | -8.454330782 |
| 246 | 0.451267185 | 0.980071786 | 0.811323051 | 5.502159445 | 1.181897373 | 9.946908549 |
| 300 | 0.669224292 | 0.749757275 | 0.479774682 | 7.225507263 | -1.117029476 | -9.036385084 |
| 304 | 0.466258133 | 0.865307244 | 0.450952233 | 1.655526875 | -1.011137978 | -2.584084229 |
| 331 | 0.808032645 | 0.79200386 | 0.862501749 | 6.434145357 | -13.20503835 | 25.47270216 |
| 332 | 0.594213493 | 0.821722198 | 0.432383986 | 3.157641943 | -3.266468152 | -5.942313014 |
| **Mean** | **0.635244894** | **0.752933323** | **0.493454399** | **7.894556226** | **-1.566812804** | **-3.748385575** |

| Category 2 SDF Ava | | | | | | |
|---|---|---|---|---|---|---|
| | Velocity ($ms^{-1}$) | | | Accleration ($ms^{-2}$) | | |
| **clip** | **Takeoff** | **Flying** | **Landing** | **Takeoff** | **Flying** | **Landing** |
| 36 | 0.433486378 | 0.801629796 | 0.649795194 | 12.99158674 | -0.749057378 | 6.921146917 |
| 43 | 0.601279332 | 1.017631625 | 0.800236547 | 0.721856554 | -2.149813147 | 2.270197721 |
| 50 | 0.669795773 | 0.816690526 | 0.324605658 | -3.885401075 | -1.000774775 | -8.263830443 |
| 61 | 0.560969714 | 0.733155475 | 0.196027025 | 6.625702417 | -17.20767819 | -8.422992496 |
| 100 | 0.838661488 | 0.756444355 | 0.87354941 | 0.24625192 | -3.694443 | 12.02516685 |
| 104 | 0.272083772 | 0.409282237 | 0.403802244 | 0.151463969 | 3.223333814 | -4.403897783 |
| 147 | 0.960280736 | 1.014953712 | 0.24052953 | 6.45374122 | 3.050463401 | -3.475673689 |
| 160 | 0.648395754 | 0.73641217 | 0.568621503 | 8.734229417 | 0.233021262 | -12.17384475 |
| 177 | 0.543940946 | 1.147590634 | 0.670235231 | -6.660131031 | 12.9876731 | -8.801958949 |
| 182 | 2.528239097 | 0.69910982 | 1.888226005 | 59.74334371 | -69.78235247 | 67.90607407 |
| 206 | 0.550527737 | 0.591743166 | 0.420713586 | 4.07730039 | -1.408987487 | 5.772025834 |
| 265 | 0.717518344 | 0.702045791 | 0.401953338 | 10.97053057 | -2.087923005 | -5.897981914 |
| **Mean** | **0.777098256** | **0.785557442** | **0.61985794** | **8.347539567** | **-6.548878157** | **3.621202614** |

| Category 2 SDF Joe | | | | | | |
|---|---|---|---|---|---|---|
| | Velocity ($ms^{-1}$) | | | Accleration ($ms^{-2}$) | | |
| **clip** | **Takeoff** | **Flying** | **Landing** | **Takeoff** | **Flying** | **Landing** |
| 5 | 1.065476405 | 0.640887677 | 0.428543638 | 15.72702964 | -6.453896382 | -12.7556883 |
| 7 | 0.409192424 | 0.924833466 | 0.171242249 | 1.688334507 | 4.451240175 | -7.41290308 |
| 9 | 0.355719908 | 1.308083267 | -0.489264259 | -3.278207911 | 21.31536606 | -56.70675626 |
| 19 | 0.733080367 | 0.759099235 | 0.149311168 | 10.62417651 | -0.963055301 | -3.142408829 |
| 37 | 0.543723209 | 0.909762415 | 0.447131967 | 6.999866011 | 6.20693646 | -29.64082409 |
| 39 | 0.650898962 | 0.467243139 | 0.407506807 | 9.839464888 | -9.316151861 | 10.6570519 |
| 73 | 0.365669032 | 1.019227336 | 0.394119634 | 3.012417046 | 13.5331058 | -29.59207194 |
| 79 | 0.37478483 | 0.5212313 | 0.326938411 | 5.831566192 | -2.567837714 | -6.188321124 |
| 87 | 0.799371359 | 0.99901243 | 0.444474753 | 8.649692502 | 2.953841152 | -9.632389228 |
| 97 | 0.585041516 | 0.652867749 | 0.618023677 | 4.489808648 | -0.952738835 | 2.511684803 |
| 121 | 0.37002329 | 0.652882176 | 0.36547898 | 0.187374392 | -0.008693734 | -5.56353493 |
| 124 | 0.448934376 | 0.620855759 | 0.716145859 | 0.112040149 | -1.303209358 | 13.88951453 |
| **125** | 0.352012368 | 0.539390368 | 0.366945897 | 0.231091598 | -3.816136771 | -3.023512087 |
| 126 | 0.224222844 | 0.325313161 | 0.430380719 | 3.584432184 | -3.118018856 | -1.725674244 |
| 228 | 1.201682462 | 0.504367327 | 0.480710836 | 23.08846663 | -2.985215725 | 0.364633634 |
| **Mean** | **0.565322223** | **0.723003787** | **0.350512689** | **6.052503533** | **1.131702341** | **-9.197413283** |

| Category 2 ODF Nigel | | | | | | |
|---|---|---|---|---|---|---|
| | Velocity ($ms^{-1}$) | | | Accleration ($ms^{-2}$) | | |
| **clip** | **Takeoff** | **Flying** | **Landing** | **Takeoff** | **Flying** | **Landing** |
| 64 | 0.376973297 | 0.786111259 | 0.142872019 | 5.768460749 | 4.936786571 | -21.41084596 |
| 137 | 0.18184904 | 0.626206936 | 0.197968735 | 1.815153051 | 3.768983078 | -6.563335639 |
| 159 | 0.672757649 | 0.741091508 | 0.538951442 | 9.234702501 | -0.63411884 | -1.848274681 |
| 166 | 0.345375595 | 0.533910822 | 0.328876488 | 4.1333773 | -0.517725264 | -1.722995898 |
| 171 | 0.355604077 | 0.979004574 | 0.545115021 | 2.449790763 | 0.524551729 | -35.02196367 |
| 178 | 1.865852656 | 0.746047108 | 0.3089373 | 44.21174948 | -29.97665225 | -14.26797159 |
| 215 | 1.335503129 | 0.605028062 | 0.767999512 | 40.02502877 | -8.246033034 | 21.1187656 |
| 237 | 0.381408843 | 0.740083474 | 1.172853559 | -0.01850261 | -1.508983574 | 26.94119833 |
| 247 | 0.83896132 | 0.859341266 | 1.763231076 | 18.47301074 | -30.98713432 | 52.87891643 |
| 295 | 0.512756165 | 0.706564296 | 1.387546713 | 6.306679123 | -15.37156559 | 53.65316192 |
| 316 | 0.355880719 | 0.727422259 | 0.440455995 | 4.990342734 | -0.504867181 | 2.132385597 |
| 318 | 0.312798032 | 0.729397126 | 0.455959174 | 3.533384641 | -1.181124609 | 5.930518419 |
| **Mean** | **0.62797671** | **0.731684057** | **0.670897253** | **11.7435981** | **-6.641490274** | **6.818296571** |

| Category 2 ODF Hedwig | | | | | | |
|---|---|---|---|---|---|---|
| | Velocity ($ms^{-1}$) | | | Accleration ($ms^{-2}$) | | |
| **clip** | **Takeoff** | **Flying** | **Landing** | **Takeoff** | **Flying** | **Landing** |
| 59 | 0.557910097 | 0.6622934 | 1.527843837 | 5.588221965 | -14.60778142 | 51.17447969 |
| 115 | 0.178091112 | 0.644502926 | 0.315150956 | 0.557963987 | 1.05992076 | -3.578723242 |
| 118 | 0.692201308 | 0.907221419 | 0.652005267 | 6.376404276 | -2.332940024 | 6.266090188 |
| 249 | 0.478086575 | 0.946533562 | 0.834047958 | -6.668341024 | 0.299792961 | 11.24365872 |
| 272 | 0.378572069 | 0.305291576 | 0.6929274 | 0.341542758 | -3.314388305 | 12.10099555 |
| 279 | 0.334944352 | 0.729635149 | 0.812881947 | 4.032074118 | 0.309997611 | 12.81780404 |
| 282 | 0.230672642 | 0.686444362 | 0.52825046 | 2.812509005 | 0.113571068 | -5.668296977 |
| 289 | 0.631134232 | 0.984995108 | 1.815297558 | 10.5706394 | -20.94641843 | 51.76687066 |
| 292 | 0.324531885 | 0.721944714 | 0.443723426 | 3.7379916 | 0.221000177 | -5.096752451 |
| 293 | 0.31095636 | 0.903656166 | 0.484534455 | 1.752941346 | 1.034022999 | 2.258243338 |
| 299 | 0.189161513 | 0.737886196 | 1.042232047 | 2.136899631 | -9.571092831 | 30.96896815 |
| 313 | 0.263502975 | 0.921639583 | 0.523365746 | 1.145214693 | 0.622577443 | -1.84092483 |
| 314 | 0.620834323 | 0.838788442 | 0.216853568 | 9.077646157 | 2.434032298 | -5.057282299 |
| 320 | 0.445323224 | 0.934982149 | 0.44558436 | 0.230063902 | 1.755488787 | -8.402181546 |
| 341 | 0.328587327 | 0.885714262 | 0.407883377 | 5.929259147 | -1.477256489 | 5.129762294 |
| 347 | 0.294346015 | 0.777597732 | 0.230543849 | 1.122486946 | 1.463678898 | -0.89375021 |
| 348 | 0.191534458 | 0.867525477 | 0.534175023 | 0.696400954 | -1.219808271 | 0.98292877 |
| 349 | 0.571289361 | 0.926527867 | 0.370226138 | 2.980192824 | 2.132069233 | -3.93910438 |
| **Mean** | **0.390093324** | **0.799065561** | **0.659862632** | **2.912228427** | **-2.334640752** | **8.346265859** |

| Category 2 ODF Ava | | | | | | |
|---|---|---|---|---|---|---|
| | Velocity ($ms^{-1}$) | | | Accleration ($ms^{-2}$) | | |
| **clip** | **Takeoff** | **Flying** | **Landing** | **Takeoff** | **Flying** | **Landing** |
| 69 | 0.669472972 | 0.711748871 | 0.517300349 | 7.039357417 | -5.346676448 | 11.25512631 |
| 70 | 0.634359297 | 1.048908791 | 0.420192306 | 3.703401986 | -1.358889513 | -11.88551443 |
| 139 | 0.415263347 | 0.510675864 | 0.081748343 | -0.018703792 | 8.373813435 | -11.43248796 |
| 150 | 0.536504072 | 0.437197606 | 0.518823406 | 5.359686999 | -0.374887942 | 4.771443485 |
| 167 | 0.236224583 | 0.713960867 | 0.252255754 | 2.024433333 | 0.716145349 | -4.495512775 |
| 191 | 0.209820512 | 0.753181247 | 0.910698902 | -4.599611355 | -2.802659702 | 6.215042829 |
| 196 | 0.339948306 | 0.480693842 | 0.659197572 | 2.400893297 | -4.148566637 | -5.605700342 |
| 201 | 0.528720474 | 0.631038356 | 0.63159735 | 8.637824987 | -1.892195449 | 15.26509323 |
| 202 | 0.328746453 | 0.508086553 | 1.049285301 | 9.852531207 | -4.689333057 | 33.44382411 |
| 209 | 0.21962413 | 0.467152398 | 0.325659736 | -0.017647271 | -6.201408614 | -4.973736684 |
| 214 | 0.223107829 | 0.666701791 | 0.940990936 | 3.242305107 | -7.226768247 | 28.77604619 |
| 229 | 0.393188619 | 0.760674419 | 0.386726214 | -6.674274644 | 4.457375468 | -23.62492486 |
| 243 | 0.614983668 | 0.744914122 | 0.813888112 | 14.27644225 | -3.897885267 | -0.884984481 |
| 274 | 0.552051785 | 0.333248769 | 0.288315945 | 11.47429093 | 0.076741414 | -1.618897678 |
| 275 | 0.350469955 | 0.428740845 | 0.384644548 | -1.930175923 | 3.680480117 | -8.443111654 |
| 281 | 0.194925254 | 0.77864276 | 0.73213318 | 2.271403311 | 2.692550804 | 6.639368489 |
| 284 | 0.169580231 | 0.381787164 | 0.237330473 | -1.669329029 | -1.064281409 | -2.365140863 |
| **Mean** | **0.389234793** | **0.609256133** | **0.538281672** | **3.257225224** | **-1.118026218** | **1.825643112** |

| Category 2 ODF Joe | | | | | | |
|---|---|---|---|---|---|---|
| | Velocity ($ms^{-1}$) | | | Accleration ($ms^{-2}$) | | |
| **clip** | **Takeoff** | **Flying** | **Landing** | **Takeoff** | **Flying** | **Landing** |
| 123 | 0.124470885 | 0.416940317 | 0.268988241 | 0.015055991 | 2.656613148 | -5.359662338 |
| 130 | 0.537243505 | 0.601190041 | 0.143885469 | -0.04699878 | -3.250482238 | -11.75756819 |
| 161 | 0.39311991 | 0.484839697 | 0.232636921 | 7.919066177 | -1.451444314 | -3.602664302 |
| 184 | 1.209457808 | 0.972447638 | 1.006755333 | 26.15146885 | -2.341182993 | -0.790570213 |
| 190 | 0.432307859 | 0.851639857 | 1.586401406 | -2.363647221 | -10.44669522 | 47.89937833 |
| 204 | 0.460060559 | 0.78583041 | 0.448890274 | -2.796674699 | 0.417125842 | -13.95232933 |
| 207 | 0.620032411 | 0.733683352 | 0.383996416 | 13.55822903 | 1.20571506 | -2.488622204 |
| 208 | 0.582589769 | 0.555581469 | 0.235319076 | 1.38546403 | -4.261881727 | -2.016170743 |
| 232 | 0.470649764 | 0.815415667 | 0.418608972 | -4.586489267 | 6.813764755 | -0.291967845 |
| **Mean** | **0.536659163** | **0.690840939** | **0.525053568** | **4.359497124** | **-1.184274188** | **0.84886924** |

| Category 2 S Nigel | | | | | | |
|---|---|---|---|---|---|---|
| | Velocity ($ms^{-1}$) | | | Accleration ($ms^{-2}$) | | |
| **clip** | **Takeoff** | **Flying** | **Landing** | **Takeoff** | **Flying** | **Landing** |
| 53 | 0.903814646 | 0.93305669 | 7.131011996 | 13.8445067 | -35.40601872 | 295.7615987 |
| 60 | 0.543992747 | 0.919260611 | 0.291901603 | 0.457337169 | 6.943061764 | -17.98872393 |
| 72 | 0.572719592 | 0.753625117 | 0.3034762 | 5.493022022 | -0.281567462 | -17.27769161 |
| 105 | 0.439387079 | 0.746833891 | 0.539810842 | -0.803492055 | 1.959057977 | 4.598202737 |
| 108 | 0.82188357 | 0.624045167 | 0.480949985 | 21.05081938 | -0.915374655 | -1.788173749 |
| 133 | 0.467363822 | 0.822670995 | 0.277345218 | 0.052526131 | 11.26422447 | -11.0808412 |
| 136 | 0.70027354 | 0.58424887 | 0.417412387 | 10.03420307 | -3.912670026 | -2.585642863 |
| 194 | 1.409400199 | 0.614174037 | 0.402129159 | 42.23972395 | -6.06924567 | -11.2920154 |
| 199 | 0.891031917 | 0.857997789 | -0.230616365 | 16.01561691 | 6.161655767 | -43.28328989 |
| 203 | 0.817415771 | 0.687319759 | 1.295363299 | 3.033831709 | -1.908742276 | 25.68813698 |
| 221 | 1.97988763 | 0.576579118 | 0.28254863 | 59.33723226 | -5.805651507 | -16.51209839 |
| 231 | 0.709905513 | 0.777623468 | 0.884623964 | 5.5737816 | 5.026912948 | -6.81324599 |
| 241 | 0.555346592 | 0.47705161 | 0.22809074 | 4.0362086 | -4.642177115 | -0.578145147 |
| 291 | 0.422438956 | 0.70907366 | 0.724105307 | 1.348785707 | -3.158911151 | 13.44912896 |
| 294 | 0.349196111 | 0.972393735 | 0.220434327 | 3.541216489 | 5.806776459 | -1.952491529 |
| 296 | 0.161456446 | 0.898684815 | 0.48954261 | 4.21217977 | 1.700771185 | 1.843266492 |
| 298 | 0.53544543 | 0.75564922 | 0.366054605 | 3.231798899 | -25.05270477 | -8.46648014 |
| 301 | 0.732962228 | 0.273449218 | 0.244756927 | 15.48284369 | -11.53938157 | -5.008786578 |
| 306 | 0.388156066 | 0.595619953 | 0.029596866 | 0.924076466 | 3.68720025 | -9.183784979 |
| 307 | 0.574822452 | 0.701255021 | 0.38765043 | 8.141960167 | -2.308041052 | 4.074059612 |
| 310 | 0.240318645 | 0.878786822 | 0.480203421 | -3.178914789 | 4.538116254 | -1.50961651 |
| 311 | 0.338204265 | 0.675324974 | 0.496718186 | 0.948335889 | -2.869020128 | 8.332268736 |
| 317 | 0.445935541 | 0.689362283 | 0.430510282 | 4.077300235 | -1.767631318 | -3.903194082 |
| 319 | 0.474717258 | 0.664390311 | 0.374808989 | 5.961141988 | -3.966552903 | 9.941261934 |
| 322 | 0.277105529 | 0.83473121 | 0.434546987 | 0.074419614 | -1.090072296 | 4.109374522 |
| 333 | 0.36308544 | 0.664148471 | 0.27777611 | 2.975213309 | -0.170537151 | -0.158602271 |
| **Mean** | **0.619856423** | **0.718744493** | **0.663875104** | **8.773295188** | **-2.452943181** | **8.015941324** |

| Category 2 S Hedwig | | | | | | |
|---|---|---|---|---|---|---|
| | Velocity ($ms^{-1}$) | | | Accleration ($ms^{-2}$) | | |
| **clip** | **Takeoff** | **Flying** | **Landing** | **Takeoff** | **Flying** | **Landing** |
| 77 | 0.420080416 | 0.76523876 | 0.469834351 | -0.139583917 | 0.161913492 | -10.14444278 |
| 96 | 0.407698586 | 0.961578436 | 0.474209047 | 1.115168087 | 4.631155599 | -12.91947155 |
| 102 | 1.12585167 | 0.803389697 | 0.568911407 | 19.41546674 | -4.020267654 | 1.433271888 |
| 103 | 0.184370179 | 0.43090525 | 0.156390909 | -0.019362534 | -1.20867079 | 0.678113318 |
| 106 | 0.432119769 | 0.545889134 | 0.29252982 | 0.089688192 | -5.906401144 | -1.691566509 |
| 112 | 0.343033133 | 0.567121486 | 1.008546857 | 3.889799411 | -3.608352667 | 24.13034553 |
| 151 | 0.461011215 | 0.761096507 | 0.439095671 | 0.147829819 | 6.163472616 | -2.785817101 |
| 163 | 2.825652564 | 0.861216674 | 0.234840123 | 73.69183811 | -58.78811701 | -17.59882696 |
| 164 | 0.606213316 | 0.780907757 | 0.37986452 | 5.062122382 | -1.532267273 | -5.97143987 |
| 278 | 0.316366797 | 0.75458863 | 1.784884358 | 6.642180935 | 1.080361893 | 31.80012692 |
| 280 | 0.296747808 | 0.814010852 | 0.45576471 | -1.924302964 | 0.739388864 | -4.03253929 |
| 302 | 0.250330368 | 0.740605844 | 0.841138546 | 2.460833822 | -3.550092749 | 18.37297069 |
| 323 | 0.317400254 | 1.00520493 | 0.164784604 | 0.253148697 | 5.887500101 | -11.09913962 |
| 336 | 0.266815355 | 0.822883562 | 0.990319224 | -0.386845692 | 4.441126928 | 11.22656231 |
| **Mean** | **0.589549388** | **0.758188394** | **0.590079582** | **7.878427221** | **-3.964946414** | **1.528439069** |

| Category 2 S Ava | | | | | | |
|---|---|---|---|---|---|---|
| | Velocity ($ms^{-1}$) | | | Accleration ($ms^{-2}$) | | |
| **clip** | **Takeoff** | **Flying** | **Landing** | **Takeoff** | **Flying** | **Landing** |
| 13 | 0.234302653 | 0.974683029 | 0.095350759 | 0.23588399 | 6.814293113 | -26.15590965 |
| 46 | 0.289135686 | 1.17424498 | 0.310920636 | -6.915953116 | 11.33709112 | -9.572603865 |
| 54 | 0.870723466 | 0.537122399 | 0.526455227 | 7.898505577 | 0.20035735 | -4.817652311 |
| **Mean** | **0.464720602** | **0.895350136** | **0.310908874** | **0.406145484** | **6.117247196** | **-13.51538861** |

| Category 2 S Joe | | | | | | |
|---|---|---|---|---|---|---|
| | Velocity ($ms^{-1}$) | | | Accleration ($ms^{-2}$) | | |
| **clip** | **Takeoff** | **Flying** | **Landing** | **Takeoff** | **Flying** | **Landing** |
| 129 | 0.086915461 | 0.334550531 | 0.536927185 | 0.871849213 | -2.867826918 | 18.74121016 |
| 195 | 0.200334278 | 0.899434017 | 1.317587911 | -14.31490176 | -3.564121156 | 39.00282904 |
| 200 | 0.608668048 | 0.954844101 | 0.668501444 | 0.230895815 | 4.778284212 | -3.721645221 |
| 213 | 0.481035055 | 0.730073849 | 0.145705549 | 0.678747797 | 4.204280501 | -10.66813205 |
| **Mean** | **0.344238211** | **0.729725624** | **0.667180522** | **-3.133352233** | **0.63765416** | **10.83856548** |

| Category 3 | | |
|---|---|---|
| Speed ($ms^{-1}$) | | |
| clp | flying | landing |
| 1 | 0.687372 | 0.314062 |
| 82 | 0.614945 | 0.221954 |
| 90 | 0.734485 | 0.292885 |
| 92 | 0.906388 | 0.484353 |
| 99 | 0.621382 | 0.251452 |
| 107 | 0.424439 | 0.140307 |
| 113 | 0.561486 | 0.438416 |
| 127 | 0.489472 | 0.403138 |
| 128 | 0.576251 | 0.481564 |
| 140 | 0.960718 | 0.273436 |
| 141 | 0.360073 | 0.128576 |
| 142 | 0.763942 | 0.540973 |
| 143 | 0.7673 | 0.313585 |
| 143 | 0.7673 | 0.313585 |
| 145 | 0.731438 | 0.451921 |
| 152 | 1.219508 | 0.244502 |
| 153 | 0.574575 | 0.458914 |
| 154 | 0.619297 | 0.370249 |
| 156 | 0.40913 | 0.238779 |
| 156 | 0.40913 | 0.238779 |
| 157 | 0.486892 | 0.377313 |
| 163 | 0.699388 | 0.213486 |
| 168 | 1.049428 | 0.534549 |
| 172 | 0.578 | 0.466766 |
| 173 | 0.770985 | 0.378614 |
| 174 | 0.523398 | 0.358964 |
| 193 | 0.596089 | 0.241658 |
| 198 | 0.490761 | 0.360779 |
| 210 | 0.920871 | 0.458491 |
| 211 | 0.817096 | 0.539124 |
| 212 | 0.44426 | 0.5247 |
| 216 | 1.05796 | 0.786951 |
| 217 | 0.626003 | 0.241737 |
| 218 | 0.907038 | 0.427445 |
| 222 | 1.398146 | 0.697757 |
| 223 | 0.788812 | 0.808357 |
| 224 | 1.097707 | 0.495175 |
| 225 | 0.738133 | 0.342304 |
| 226 | 0.9942 | 0.509202 |
| 227 | 0.762411 | 0.748854 |
| 233 | 0.862126 | 0.392265 |
| 240 | 0.60834 | 0.486099 |
| 248 | 1.525215 | 0.426539 |
| 250 | 1.135578 | 0.606824 |
| 251 | 1.508399 | 0.544692 |
| 252 | 0.728441 | 0.250469 |
| 253 | 1.7944 | 0.649577 |
| 254 | 1.218678 | 0.389058 |
| 255 | 0.665797 | 0.372009 |
| 256 | 0.806709 | 0.526268 |
| 258 | 0.785734 | 0.181913 |
| 259 | 0.599353 | 0.574325 |
| 260 | 0.585542 | 0.297242 |
| 261 | 0.677392 | 0.398499 |
| 262 | 0.586249 | 0.371537 |
| 263 | 0.703038 | 0.343989 |
| 270 | 0.539329 | 0.267687 |
| 271 | 0.675285 | 0.229339 |
| 273 | 0.389443 | 0.249852 |
| 276 | 0.902561 | 0.374163 |
| 286 | 1.081792 | 0.409962 |
| 308 | 0.395372 | 0.195589 |
| 315 | 0.835917 | 0.489545 |
| 334 | 0.847831 | 0.40507 |
| 351 | 0.579677 | 0.365911 |
| 351 | 0.579677 | 0.365911 |
| 352 | 0.514017 | 0.202491 |
| 353 | 1.140642 | 0.269691 |
| 354 | 0.880458 | 0.365542 |
| **Mean** | **0.769554** | **0.392981** |